%% file: main.tex
\documentclass[dvipsnames]{article} %
\usepackage{colm2024_conference}

\usepackage{amsmath, amssymb, amsfonts, amsthm}
\let\AMSmathOverline\overline

\usepackage[utf8]{inputenc}
\usepackage[T1]{fontenc}
\usepackage{textcomp}

\usepackage{graphicx}
\usepackage{tikz}
\usetikzlibrary{er,positioning,bayesnet,tikzmark} 
\usepackage{pgfplots}
\pgfplotsset{compat=1.18}
\usepackage{wrapfig}

\usepackage{booktabs}
\usepackage{multirow}
\usepackage{colortbl}
\usepackage{array}
\usepackage{makecell}
\usepackage{xltabular}
\usepackage{tabularx}
\usepackage{threeparttable}
\usepackage{rotating}

\usepackage{tablefootnote}
\usepackage{arydshln}
\usepackage{hhline}

\usepackage{xcolor}
\definecolor{lightgray}{rgb}{0.9,0.9,0.9}
\usepackage{hyperref}
\usepackage{xurl}
\usepackage{url}

\usepackage{microtype}
\usepackage{nicefrac}
\usepackage{setspace}

\usepackage[misc]{ifsym}
\usepackage{pifont}
\usepackage{fontawesome}
\usepackage{tipa}
\usepackage{siunitx}
\usepackage{subscript}
\usepackage{mathrsfs}
\usepackage{latexsym}
\usepackage{inconsolata}

\usepackage{listings}
\usepackage{adjustbox}
\usepackage{verbatim}

\usepackage[raster,skins,most]{tcolorbox}

\usepackage{caption}
\usepackage{subcaption}

\usepackage[ruled,vlined,linesnumbered]{algorithm2e}
\SetKw{Continue}{continue}

\usepackage{enumitem}
\usepackage{blindtext}

\newtheorem{proposition}{Proposition}

\newcommand{\w}{WebLeaper}

\input{math_commands.tex}

\renewcommand{\mslink}{https://modelscope.cn/datasets/iic/WebLeaper}
\renewcommand{\ghlink}{https://github.com/Alibaba-NLP/DeepResearch}

\usepackage{makecell}
\usetikzlibrary{tikzmark}
\makeatletter
\newcommand*\myfontsize{%
  \@setfontsize\myfontsize{7}{8}%
}
\makeatother

\definecolor{uclablue}{RGB}{159, 195, 224}

\definecolor{uclagold}{RGB}{255, 240, 180}

\definecolor{aliceblue}{RGB}{255, 238, 241}

\definecolor{cadmiumgreen}{rgb}{0.0, 0.42, 0.24}

\definecolor{myred}{rgb}{0.7, 0.3, 0.0}
\definecolor{myblue}{rgb}{0.2, 0.3, 0.6}
\definecolor{babygreen}{rgb}{0.85, 0.97, 0.85}

\definecolor{purple1}{RGB}{126, 107, 196}
\definecolor{purple2}{RGB}{199, 158, 207}
\definecolor{purple3}{RGB}{214, 200, 255}
\definecolor{purple4}{RGB}{254, 240, 255}
\definecolor{custompurple}{HTML}{491f97}
\definecolor{deepblue}{RGB}{48, 58, 82}
\definecolor{darkred}{HTML}{C00000}

\newcommand{\symboletongyi}{\raisebox{0pt}{~\includegraphics[scale=0.012]{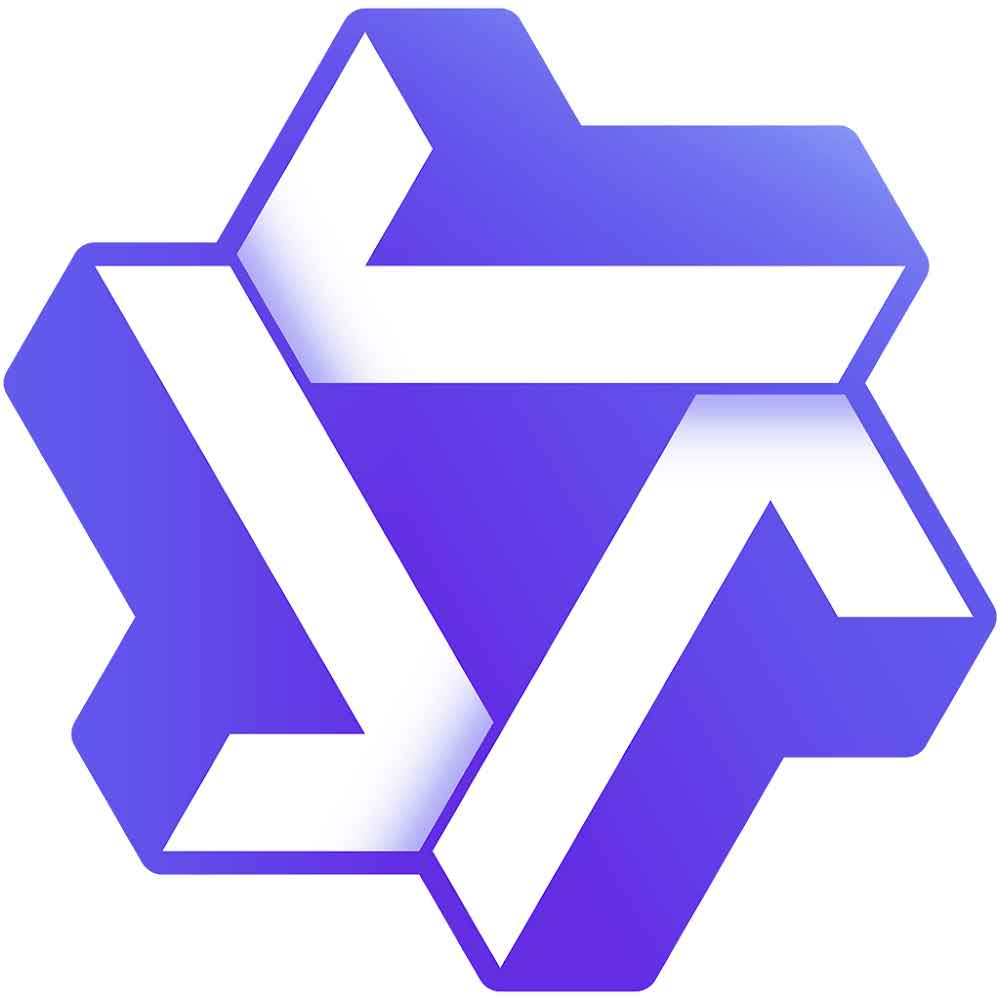}}~}

\definecolor{deepPurple}{HTML}{330066}

\definecolor{uclablue_old}{rgb}{0.15, 0.45, 0.68}
\hypersetup{
    breaklinks,
    citecolor=uclablue_old,
    colorlinks=true,
}

\newtcolorbox{mybox}[2][]
  {colback = black!5!white, colframe = black!75!black, fonttitle = \bfseries,
    colbacktitle = black!100!black, enhanced, before upper={\fontsize{8}{11}\obeyspaces\obeylines\selectfont}, fontupper=\selectfont,
    attach boxed title to top left={yshift=-2.2mm,xshift=4mm},
    title=#2,#1}

\let\overline\AMSmathOverline

\title{%
\raisebox{-2.3em}{
  \parbox[t]{0.35in}{\includegraphics[width=0.6in]{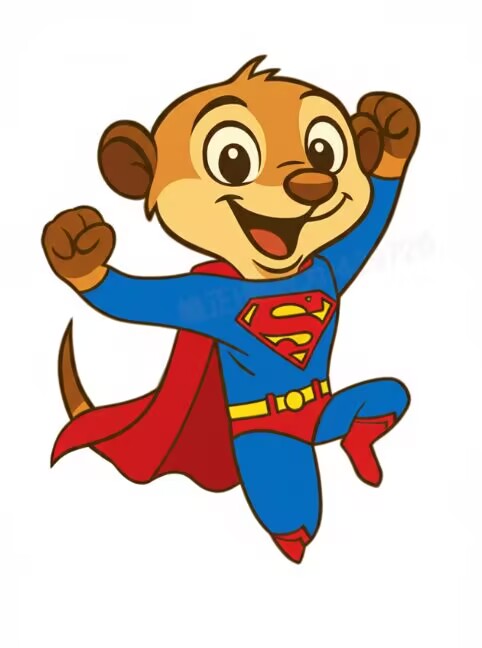}} 
}
\begin{tabular}[t]{l} 
  \parbox[t]{0.8\textwidth}{\centering 
    \w: Empowering Efficiency and Efficacy in WebAgent via Enabling Info-Rich Seeking
  }
\end{tabular}
}

\author{%
\small{Zhengwei Tao$^{*}$$^{(\textrm{\Letter})}$, Haiyang Shen$^{*}$, Baixuan Li$^{*}$, Wenbiao Yin$^{(\textrm{\Letter})}$, Jialong Wu, Kuan Li, Zhongwang Zhang, Huifeng Yin, Rui Ye, Liwen Zhang, Xinyu Wang, Pengjun Xie, Jingren Zhou, Yong Jiang$^{(\textrm{\Letter})}$}%
  \\[1em]               
  {\fontsize{10pt}{11pt}\selectfont          
Tongyi Lab\symboletongyi, Alibaba Group}\\
}

\begin{document}

\maketitle

\begingroup
  \renewcommand\thefootnote{*} 
   \footnotetext{Equal contribution. }
\endgroup

\begingroup
\renewcommand\thefootnote{$^{\textrm{\Letter}}$}
\footnotetext{Correspondence to: \texttt{tttzw@stu.pku.edu.cn}, \texttt{\{yinwenbiao.ywb, yongjiang.jy\}@alibaba-inc.com}.}
\endgroup


\begin{abstract}

Large Language Model (LLM)-based agents have emerged as a transformative approach for open-ended problem solving, with information seeking (IS) being a core capability that enables autonomous reasoning and decision-making. 
While prior research has largely focused on improving retrieval depth, we observe that current IS agents often suffer from \textit{low search efficiency}, which in turn constrains overall performance. 
A key factor underlying this inefficiency is the sparsity of target entities in training tasks, which limits opportunities for agents to learn and generalize efficient search behaviors.
To address these challenges, we propose \w, a framework for constructing high-coverage IS tasks and generating efficient solution trajectories.
We formulate IS as a tree-structured reasoning problem, enabling a substantially larger set of target entities to be embedded within a constrained context. 
Leveraging curated Wikipedia tables, we propose three variants for synthesizing IS tasks—\texttt{Basic}, \texttt{Union}, and \texttt{Reverse-Union}—to systematically increase both IS efficiency and efficacy. Finally, we curate training trajectories by retaining only those that are simultaneously accurate and efficient, ensuring that the model is optimized for both correctness and search performance.
Extensive experiments on both basic and comprehensive settings, conducted on five IS benchmarks—BrowserComp, GAIA, xbench-DeepSearch, WideSearch, and Seal-0—demonstrate that our method consistently achieves improvements in both effectiveness and efficiency over strong baselines.

\end{abstract}





\input{sections/1-introduction}

\input{sections/2-definition}
\input{sections/3-method}
\input{sections/4-experiments}
\input{sections/5-related_work}
\input{sections/6-conclusion}

\clearpage
\bibliography{biblio}
\bibliographystyle{colm2024_conference}

\clearpage
\appendix
\input{sections/appendix}

\end{document}

%% file: math_commands.tex
\usepackage{amsmath,amsfonts,bm}

\def\eqref#1{equation~\ref{#1}}

\def\1{\bm{1}}

\DeclareMathAlphabet{\mathsfit}{\encodingdefault}{\sfdefault}{m}{sl}
\SetMathAlphabet{\mathsfit}{bold}{\encodingdefault}{\sfdefault}{bx}{n}

\def\gH{{\mathcal{H}}}

\def\gP{{\mathcal{P}}}

\def\gR{{\mathcal{R}}}

\def\gT{{\mathcal{T}}}



%% file: sections/1-introduction.tex
\begin{figure*}[h]
\setlength{\abovecaptionskip}{-2pt}   
\setlength{\belowcaptionskip}{-2pt}  
    \centering
    \includegraphics[width=1\textwidth]{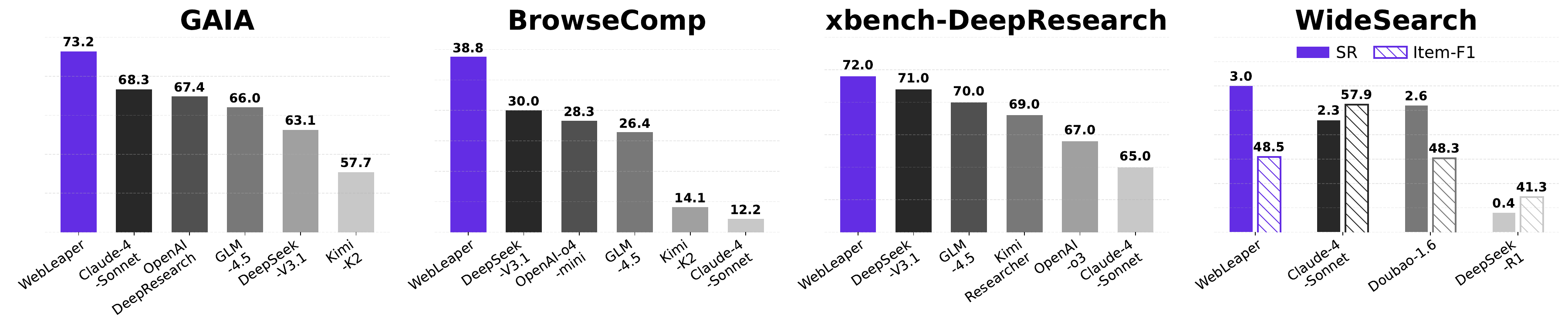}
    \caption{Results on comprehensive training setting. All  \w~scores are averaged over three runs. The metric of the first three figures are accuracy. ``SR'' denotes Success Rate on WideSearch.}
    \label{fig:main_results}
    \vspace{-1em}
\end{figure*}

\section{Introduction}
\label{sec:introduction}

The LLM-based agents mark a paradigm shift in AI, delivering transformative solutions to challenges once deemed intractable across diverse domains~\citep{guo2024large, mplug-owl}.
Among their core capabilities, information seeking (IS) plays a crucial role in enabling the cognitive autonomy of these agents. This ability not only drives their adaptability in open-ended tasks but also underpins a new generation of powerful commercial systems, including OpenAI Deep Research~\citep{openaidr}, Google’s Gemini~\citep{geminidr}, and Perplexity AI~\citep{perplexity}, Kimi-Researcher~\citep{kimi_researcher}.

\begin{wrapfigure}{r}{0.5\linewidth}
\vspace{-3em} 
    \centering
    \includegraphics[width=0.99\linewidth]{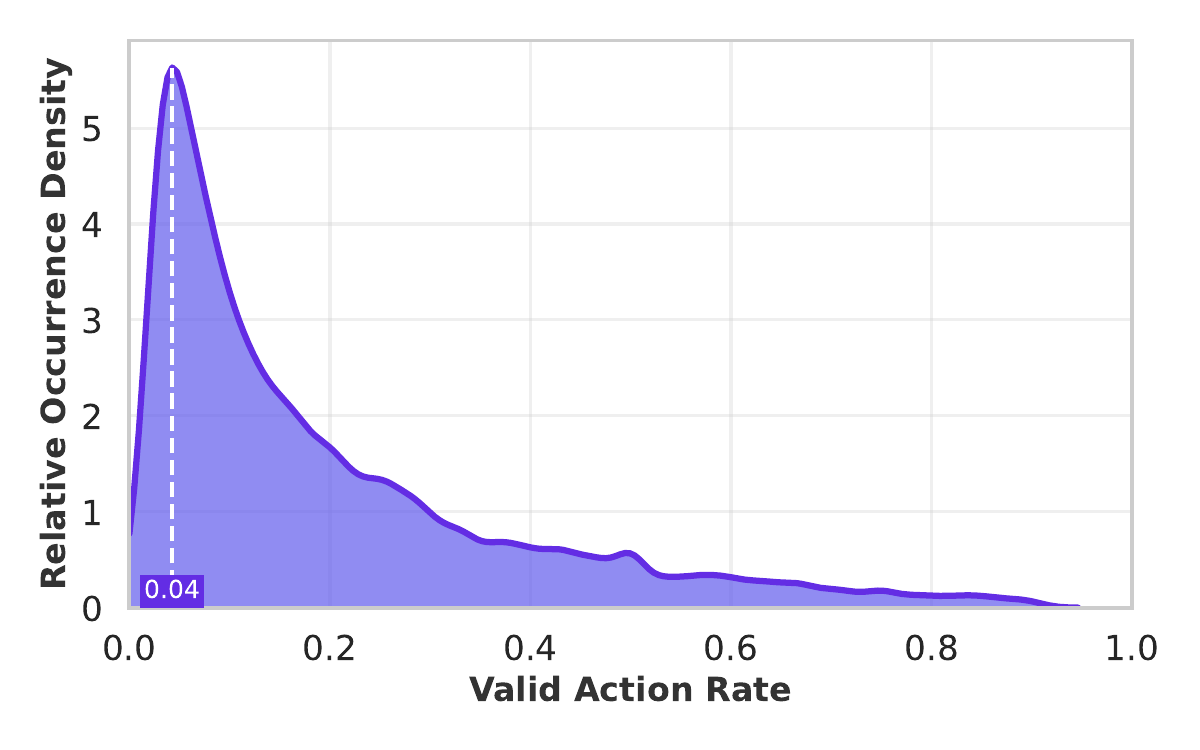}
    \caption{The distribution of valid actions of the agent based on the GPT model on our synthesized IS task. The valid actions are those seeking the correct target entities required by the question.}
    \label{fig:infogain_cover_rate_kde}
\end{wrapfigure}

While numerous studies have sought to enhance the IS capabilities of agents through complex question–answering pipelines and advanced fine-tuning strategies~\citep{wu2025webdancerautonomousinformationseeking,li2025websailornavigatingsuperhumanreasoning,li2025websailorv2bridgingchasmproprietary,tao2025webshaper,qiao2025webresearcher,lu2025deepdive}, most existing approaches primarily concentrate on improving the search depth, giving comparatively little attention to search efficiency.
Our preliminary experiments indicate that current LLM-based agents search inefficiently.
As shown in Figure~\ref{fig:infogain_cover_rate_kde}, the distribution of valid actions for a competitive IS agent peaks around 0.04, meaning that in most cases, only a small fraction of actions are effective~\citep{wong2025widesearch, xue2025simpletir}.
This low valid-action rate reflects suboptimal search behaviors, including redundant query reformulations, retrieval of irrelevant information, and unnecessarily long search chains.
Such inefficiencies not only increase computational and time costs but also limit the agent’s overall IS performance.

The design of synthetic training tasks incurs this inefficiency. In typical IS agent setups, the agent begins with a set of known entities and incrementally gathers information to infer all target entities. However, prior work often constructs tasks in which the target entities are overly sparse~\citep{wu2025webdancerautonomousinformationseeking,li2025websailornavigatingsuperhumanreasoning,li2025websailorv2bridgingchasmproprietary}. 
Such sparsity limits the agent’s exposure to informative cues, reducing opportunities to learn to locate relevant information within a constrained context window. As a result, the agent spends more actions processing irrelevant content, weakening its search strategies, leading to lower performance.
Furthermore, it can bias the measurement of search efficiency, which we prove in a later section. This bias makes it difficult to obtain an accurate training signal, thereby obstructing the systematic learning of more efficient search behaviors. These limitations underscore the need to redesign training tasks, enabling optimized seeking efficiency and stronger IS capabilities.

To address these challenges, we propose \w, a framework designed with two core objectives: (1) to construct new IS tasks containing a substantially larger number of target entities; and (2) to generate solution trajectories that achieve both high accuracy and high efficiency. For the first objective, we model the IS process as a tree-structured reasoning task, which compactly accommodates more target nodes within a limited context. Based on this formulation, we systematically increase task complexity through three dataset variants. First, leveraging curated Wikipedia tables, we synthesize \underline{\textbf{\texttt{Basic}}} version, which directly addresses the challenge of entity sparsity by creating a high-density search space within a single, structured source. To mirror more realistic scenarios that demand integrating information from multiple sources, our \underline{\textbf{\texttt{Union}}} variant constructs tasks that require synthesizing facts across different sources, thereby increasing search ambiguity. Finally, to mitigate the risk of agents adopting simplistic, keyword-based shortcuts, the \underline{\textbf{\texttt{Reverse-Union}}} variant reverses the logical flow, compelling the agent to first deduce intermediate entities from scattered clues before completing the main search task.
For the second objective, we construct task-completion trajectories that are filtered according to \emph{Information-Seeking Rate} (ISR) and \emph{Information-Seeking Efficiency} (ISE), retaining only those that solve the task both accurately and efficiently. 
These metrics are then incorporated into the hybrid reward system during the following reinforcement learning stage.
Models trained on this curated dataset after supervised-finetuning and reinforcement learning yield our final IS agent.

We conduct extensive experiments on both basic and comprehensive settings to evaluate our approach across five benchmarks: BrowserComp~\citep{bc_en}, GAIA~\citep{mialon2023gaia}, Seal-0~\citep{pham2025sealqa}, WideSearch~\citep{wong2025widesearch}, and xbench-DeepSearch~\citep{xbench}. Our method achieves consistent improvements on all benchmarks. Ablation studies on the dataset design further confirm the effectiveness of our proposed components. We summarize our contribution as follows:

\vspace{-5pt}
\begin{itemize}[left=0.2cm]\setlength{\itemsep}{0pt}\setlength{\parskip}{0pt}\setlength{\topsep}{-5pt}
    \item  We design a new information-seeking task formulation on a tree-structured reasoning problem, leading to the inclusion of a substantially larger set of target entities within a constrained context. Based on this formulation, we construct the \emph{Basic}, \emph{Union}, and \emph{Reverse-Union} datasets.
    \item We generate and filter task-solving trajectories using the proposed \emph{Information-Seeking Rate} (ISR) and \emph{Information-Seeking Efficiency} (ISE) metrics, retaining only those trajectories that solve tasks both accurately and efficiently. These metrics are also designed for our hybrid RL reward system.
    \item  We conduct extensive experiments on five public IS benchmarks, BrowserComp, GAIA, Xbench-DeepSearch, WideSearch, and Seal-0, achieving consistent improvements over strong baselines. 
\end{itemize}

%% file: sections/2-definition.tex
\section{Definitions}
\label{sec:definitions}

An Information-Seeking (IS) task challenges an agent to answer a complex natural language question by navigating a vast information space to assemble a complete set of required entities. This process is inherently sequential, involving the progressive discovery of entities, understanding their properties (attributes), and leveraging relationships between them to uncover further entities. This section formally defines the components of such a task and the metrics for evaluating an agent's performance, emphasizing the importance of identifying both final and intermediate entities in the reasoning chain.

\subsection{Information-Seeking Task}
\label{subsec:IS_task}

An entity $e\in\mathcal{E}$ is the fundamental unit of information. 
An \emph{Information-Seeking (IS) task} is the process of identifying and collecting a specific set of target entities from $\mathcal{E}$, based on a question. Formally, an IS task is a tuple:
\(\mathcal{T} = \langle q, R \rangle \)
, where $q$ is the natural language question and $R \subset \mathcal{E}$ is the set of the target entities that collectively satisfy the conditions posed by $q$.

Critically, the required set $R$ includes not only the final, explicit answers but also all \emph{intermediate entities} that are necessary stepping stones in the reasoning process. Consider the question:
\begin{equation}
\label{eq:example_IS}
\begin{aligned}
q: & \ \textit{Which player of a team in the 2004--05 season, who was born in the 1990s?} \\
   & \ \textit{This team was founded in 1966 and is an East German football team.}
\end{aligned}
\end{equation}
To solve this, an IS agent must seek for information online, and find the target entity set as answer:
\begin{equation}
\label{eq:set_R}
\begin{aligned}
R = \{ \textit{Robert Rudwaleit}, \textit{Danny Kukulies}, \ldots \}.
\end{aligned}
\end{equation}



\subsection{Information-Seeking Agent}
\label{subsec:IS_agent}

We focus on an \emph{Information-Seeking Agent} that interacts with a web environment to solve an IS task $\mathcal{T}$ within the ReAct framework~\citep{yao2023react}. The agent's operation is a sequential decision-making process occurring over discrete time steps $t=1, \dots, T$. At each step, the agent analyzes its current state (including the initial question and all previously gathered information), generates a thought for planning its next move, executes a tool-based action to seek new information, and receives an observation from the environment. This entire process is captured in the \emph{agent trajectory} is defined as
\begin{equation}
\label{eq: trajectory}
\mathcal{H}_T = (q, \tau_1, \alpha_1, o_1, \tau_2, \alpha_2, o_2, \ldots, \tau_{T}, \alpha_{T}, o_{T}),
\end{equation}
where $\tau_i$ is the planning thought, $\alpha_i$ is the seeking action, and $o_i$ is the resulting observation at step $i$. 
At the end of the process, the agent has obtained a set of entities $O \subset \mathcal{E}$, which is the union of all unique entities discovered across all steps.


\subsection{Quantifying Information Collection and Efficiency}
\label{subsec:quantifying_collection}

To guide an agent towards successfully solving IS tasks, its performance framework must value the entire reasoning process, not merely the final output. 
Our central thesis is that by explicitly quantifying the value of \emph{all} required information discovered, we can create a stronger signal for learning effective search strategies. To this end, we define principles to formalize the performance (the total information gain) and the efficiency (the gain per action) of the agent's collection process.

\vspace{-2mm}
\paragraph{Information-Seeking Rate (ISR)}
Recall that $R$ denotes the set of target ground-truth entities for the task, with cardinality $n = |R|$. $O$ is the set of entities actually obtained by the agent during its operation.  
The intersection $R \cap O$ therefore contains all required entities that were successfully retrieved.
The \emph{information collection rate} directly measures the fraction of required entities successfully obtained by the agent:
\begin{equation}
\label{eq:ISR}
\mathrm{ISR} = \frac{|R \cap O|}{|R|} = \frac{|R \cap O|}{n}.
\end{equation}
$\mathrm{ISR} \in [0,1]$, and higher values indicate more thorough coverage of the required information.  
\vspace{-2mm}
\paragraph{Information-Seeking Efficiency (ISE)}
While $\mathrm{ISR}$ measures completeness, the \emph{information collection efficiency} reflects the average number of action steps to discover the target entity:
\begin{equation}
\label{eq:ISE}
\mathrm{ISE} = \frac{n}{T},
\end{equation}

where $T$ is the total number of steps of the solving trajectory. Higher $\mathrm{ISE}$ implies greater IS efficiency. The stability of measuring $\mathrm{ISE}$ is important for providing unbiased training signals.

\begin{proposition}[Variance of ISE]
\label{prop:var_reduction}
Let $X_i$ denote the number of steps the agent takes to discover the $i$-th new entity in $R$. Therefore $\mathrm{ISE} = \frac{n}{T}=\frac{n}{\sum_{i=1}^n X_i}$.
Assume $X_1,\dots,X_n$ be i.i.d.\ random variables with finite mean $\mu>0$ and finite variance $\sigma^2$, $X_i>0$ almost surely, then:

\begin{align}
\label{eq: V}
\mathrm{Var}(\mathrm{ISE}) = \mathcal{O}\left(\frac{1}{n}\right).
\end{align}
\vspace{-3mm}

\end{proposition}

This proposition shows that as the number of target entities $n$ grows, measuring $\mathrm{ISE}$ becomes a more stable and reliable performance metric. The detailed proof is provided in Appendix~\ref{sec:appendix_proof_ise}.

%% file: sections/3-method.tex
\section{Method}
\label{sec:method}

To enhance the information efficiency of the IS agent, our approach trains the model on a calibrated task
$\gT = \langle q, R \rangle$ together with the corresponding task-solving trajectory $\gH$.
In prior IS agent training setups, the dataset typically contained only a limited number of target entities ($R$). 
This design substantially restricts the potential improvement in information-seeking efficiency and, in turn, 
limits the agent's overall capability. The limitation incurs two problems:
\vspace{-0.5em}
\begin{itemize}[left=0.2cm,itemsep=-1pt]
    \item With a small volume of $R$, it is difficult to train the agent to retrieve information efficiently within a limited context length.
    \item Our method relies on measuring the information-seeking efficiency ISE. As shown in Eq.~(\ref{eq: V}), a small set of target entities introduces measurement bias in the ISE metric.
\end{itemize}

To overcome these shortcomings, we introduce \w, a novel data synthesis framework specifically designed to 
boost information-seeking efficiency. 
Our method consists of two main components: (1) a QA synthesis pipeline for generating calibrated tasks, and (2) 
a trajectory construction process for producing realistic task-solving sequences. 
We describe the QA synthesis pipeline and trajectory construction process in detail in the following subsections. For detailed walkthroughs of the examples for each synthesis version, please refer to Appendix~\ref{sec:appendix_examples}.

\subsection{Entity-Intensive Task Synthesis}

\begin{figure}
\centering
\setlength{\belowcaptionskip}{-1mm}

\includegraphics[width=1\linewidth]{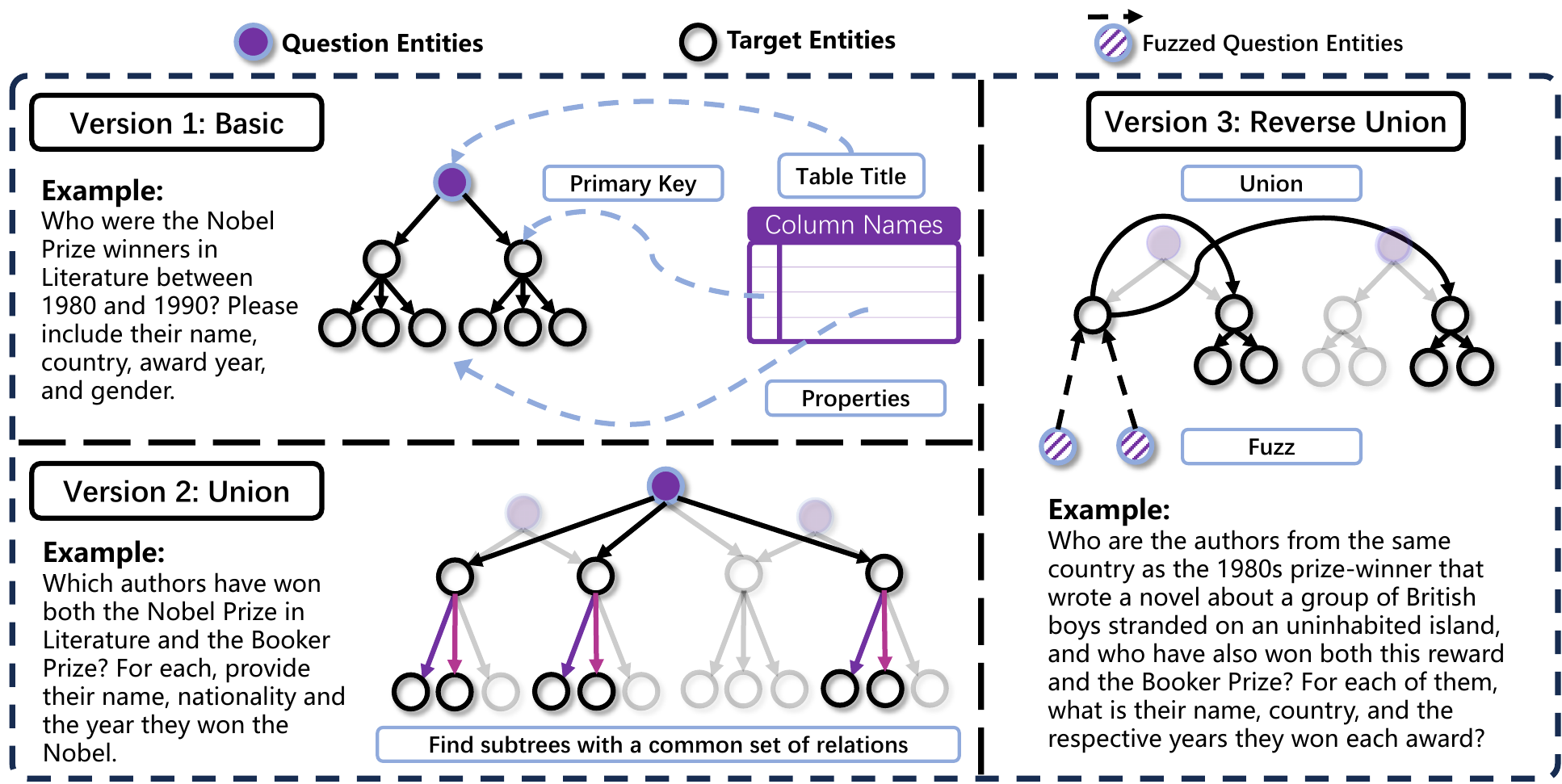}
\caption{An overview of \w{}. The reasoning structure is modeled as a tree. A root entity (question entity) connects to a set of second-layer entities. 
\textbf{(a) Version-I (Basic)} constructs a simple reasoning tree from a single information source.
\textbf{(b) Version-II (Union)} creates a complex task by finding a maximal union between two trees that share a common set of relations within their subtrees (e.g., both have ``has\_nationality''). 
\textbf{(c) Version-III (Reverse-Union)} reverses the reasoning process. It provides fuzzed clues (third-layer entities) as question entities, forcing the agent to first deduce a second-layer anchor entity (an entity from the second layer), then other relevant subtrees.}
\label{fig:overview}
\end{figure}

\subsubsection{Version-I: \texttt{Basic}}
\label{sec:base_dataset_construction}

In an information-seeking task, the reasoning structure matters. We use a tree, denoted as $T_i$, to represent this structure, where nodes are entities and edges are relations between them. The IS agent must start with some known entities in the tree and reason along the edges to determine the target ones. To incorporate as many target entities as possible, we use this tree structure for its compact and hierarchical organization.

Synthesizing such a task $\gT = \langle q, R \rangle$ requires a large volume of relevant entities, which is non-trivial. Following the one-entity-at-a-time collection strategy of prior work is prohibitively expensive. Therefore, we exploit the structured tables contained in Wikipedia articles, which encapsulate rich relational information. These tables naturally provide groups of entities connected by specific relationships, enabling us to efficiently construct the reasoning tree $T_i$. We crawled approximately 2 million tables from Wikipedia and applied a multi-stage cleaning procedure, retaining only large, well-formed, and structurally homogeneous tables. The detailed data cleaning procedure and construction rationale are described in Appendix~\ref{sec:appendix_cleaning_basic}.

To construct the reasoning structure illustrated in Figure~\ref{fig:overview}(a), we populate its layers using information from a single table. 
The entities extracted from the table title form the root of the tree (i.e., the question entities).  
Next, we employ an LLM to select the most representative, non-redundant column of values from the table—typically the primary key—as the second-layer entities (e.g., ``Czesław Miłosz'').  
An edge between the root entity and a second-layer entity indicates that the table contains this entity.  
The third-layer entities are derived from the remaining columns of the table, with their values representing attributes of the corresponding second-layer entity (e.g., ``country: Poland'', ``year: 1980'').  
In this layer, an edge signifies that the second-layer entity possesses the given property defined by the third-layer entity.

Each second-layer entity and its associated third-layer entities form a subtree, which we denote as $S_{i,j}$. These subtrees, each possessing a set of relations $\text{Rel}(S_{i,j})$ that connect its layers, represent cohesive units of information (e.g., a specific laureate and all their details). The full reasoning tree $T_i$ is thus composed of a set of such subtrees $\{S_{i,j}\}$. The question provides the root entities, while all entities in the subtrees (both second and third layers) constitute the final answer. The detailed construction process and the required reasoning path for the example task are explained in Appendix~\ref{sec:example Version-I: Basic}.

\subsubsection{Version-II: \texttt{Union}}
\label{sec:union_synthesis}

While effective, the reasoning structure of our basic tasks is derived from single sources, limiting their structural complexity and the scope of questions we can pose. To address this, we aim to construct tasks with a more intricate reasoning structure that spans multiple information sources by uniting reasoning trees from our \texttt{Basic} version that share similar themes and structures.

To generate more challenging questions, we propose uniting reasoning subtrees in \texttt{Basic} version that share similar themes and structures. 
A naive approach, such as randomly combining subtrees, often results in semantically incoherent questions. To systematically discover the most substantial integration opportunities, our approach models this as a \texttt{Union} operation, which identifies multiple reasoning trees whose respective subtrees share some common relations. 

The primary challenge is to systematically search the entire collection of trees to find all groups that are suitable for union. To avoid a combinatorial explosion from enumerating all possible combinations, we develop an algorithm to efficiently discover only \textit{maximal unions}. This problem is formally modeled as Maximal Biclique Enumeration (see Appendix~\ref{sec:appendix_maximal_fusion}), which effectively identifies groups of reasoning subtrees and their shared subtree relations.

As illustrated in Figure~\ref{fig:overview}(b), the reasoning trees for ``Nobel Prize in Literature laureates'' and ``Booker Prize winners'' both contain subtrees where second-layer entities (authors) are connected to third-layer entities via relations like ``has\_nationality'' and ``has\_name''. Our method identifies this shared subtree structure. Relations not shared across all sets of subtrees, such as ``has\_gender'' (present only in the Nobel tree), are discarded during the union.

Once a maximal union is identified, we leverage an LLM to
synthesize a question based on the common features of the selected subtrees.
For instance, the question ``Which authors have won both the Nobel Prize in Literature and the Booker Prize?'' requires identifying the two sets of laureates as intermediate ``Target Entities'' and then finding their intersection to produce the final ``Target Entities''. The complete walkthrough is in Appendix~\ref{sec:example Version-II: Union}.

\subsubsection{Version-III: \texttt{Reverse-Union}}
\label{sec:fuzz_generation}

While the \texttt{Union} method generates complex, multi-source tasks, a vulnerability remains: an agent could solve the query and use direct keyword searches on the constituent sources (e.g., search ``Nobel Prize winners,'' then ``Booker Prize winners''). This approach circumvents the intended synthesis of information, reducing the cognitive load and failing to stimulate true reasoning capabilities similar to \texttt{WebSailor}~\citep{li2025websailornavigatingsuperhumanreasoning}.
To address this, we introduce \texttt{Reverse-Union}, a paradigm designed to enforce a more robust cognitive workflow by reversing the standard reasoning flow. As illustrated in Figure~\ref{fig:overview}(c), this method combines two stages to construct a challenging task:
\vspace{-1em}
\begin{itemize}[left=0.2cm,itemsep=-1pt]
    \item \textbf{Deductive Fuzz:} This stage implements the fuzz by defining the ``Question Entities'' as a set of descriptive third-layer entities. Instead of being named directly, a central ``anchor'' entity (an entity from the second layer) is described through its corresponding third-layer entities. In the example, the description ``the 1980s prize-winner that wrote a novel about a group of British boys stranded on an uninhabited island'' serves as clues in the form of ``Question Entities''. An agent must first deduce from these clues to identify the anchor entity, ``William Golding''.

    \item \textbf{Union-based Search Construction:} After fuzzing the anchor, this stage constructs the expansive search part of the task, ensuring the anchor serves only as a bridge to the final answer. To achieve this, we first select a specific third-layer entity from the anchor's subtree (e.g., his country) to act as a pivot. We then formulate the remainder of the question to compel an agent to use this pivot to launch a new search across the unified trees. The final Target Entities are thus defined as the set of second-layer entities that share this pivot attribute (i.e., are also British) and satisfy the original intersection condition (i.e., winning both prizes).
\end{itemize}
By structuring tasks this way, \texttt{Reverse-Union} prevents agents from succeeding with simple keyword searching and mandates a more robust, multi-step reasoning process. The detailed process of question generation and the required reasoning path are explained in Appendix~\ref{sec:example Version-III: Reverse-Union}.

\subsection{Information-Guided Trajectory Construction}
\label{sec:reject_sampling}

After synthesizing the task, this section elaborates on the construction of task-solving trajectories. 
As shown in Eq.(\ref{eq: trajectory}), our agent solves a task within the ReAct framework~\citep{yao2023react}. We equip the agent with the following tools:
\vspace{-0.5em}
\begin{itemize}[left=0.4cm,itemsep=-1pt]
    \item \texttt{Search} This action enables the agent to conduct Google search by several queries. The parameters of this tool are $\{\textit{queries}, \textit{filter\_year}\}$, enabling temporal filtering of search results. This tool would return the top relevant URLs and their snippets as the observation.
    
    \item \texttt{Visit} This action enables the agent to visit multiple URLs. The parameters of this tool are $\{\textit{urls}, \textit{goal}\}$. This tool would return the summarized visited paragraphs as the observation.
    
\end{itemize}

After generating a large set of trajectories by executing our constructed tasks with an open-source model, we apply a filtering procedure to select high-quality examples for training.
Our goal is to retain trajectories that demonstrate both accuracy in collecting the required entities and efficiency in the use of actions, in accordance with the metrics defined in Section~\ref{subsec:quantifying_collection}.
Specifically, we impose the following selection criteria:


\paragraph{Coverage Criterion.}
We require that the trajectory achieve sufficient completeness in information collection. Formally, we keep only those trajectories whose ISR satisfies $\mathrm{ISR} > \alpha$, where $\alpha$ is a predefined coverage threshold. To compute ISR, we accumulate the obtained target entities in all actions. We compute $\mathrm{ISR}$ as Eq.(~\ref{eq:ISR}).


\paragraph{Efficiency Criterion.}
We further require that the trajectory maintain high efficiency in discovering useful entities. This translates into selecting those trajectories whose ISE satisfies $\mathrm{ISE} > \beta$, where $\beta$ is a predefined efficiency threshold. For $\mathrm{ISE}$, we accumulate the obtained target entities in \texttt{Visit} actions. The reason for not including $\texttt{Search}$ in $\mathrm{ISE}$ is that we observe entities found in $\texttt{Search}$ are less precise and would be updated by the following $\texttt{Visit}$ action. We compute $\mathrm{ISR}$ as Eq.(\ref{eq:ISE}).

Through this filtering process, we ensure that the retained trajectories are both accurate in acquiring the target entities and efficient in their action usage, providing strong supervision signals for training agents to perform precise and effective information-seeking.

\subsection{Reinforcement Learning with Hybrid Reward Systems}
\label{sec:hybrid_reward_system}

Following supervised fine-tuning (SFT) on the trajectories generated via our information-guided method (Section~\ref{sec:reject_sampling}), we further enhance the agent's policy using reinforcement learning (RL). A critical component of RL is the reward function, which provides the training signal. However, standard reward mechanisms are fundamentally misaligned with the entity-intensive tasks synthesized by \w{}. The most common approach, a simple binary reward (e.g., success/failure), suffers from extreme sparsity. This issue is dramatically exacerbated in our setting, where a task may require dozens of entities; rewarding the agent only upon perfect completion of such a large set makes positive feedback so rare that effective learning becomes nearly impossible.

Furthermore, the very methods for implementing a reward function—even a more granular one—present their own intractable challenges. On one hand, conventional automated metrics like Exact Match or word-level F1 scores are too brittle. They cannot gracefully handle minor semantic variations (e.g., ``USA'' vs. ``United States'') and would incorrectly penalize the agent, a problem that compounds severely across a large entity set. On the other hand, deploying a more sophisticated LLM-as-a-Judge to evaluate correctness seems promising, but it struggles with scalability and reliability. Asking a judge model to accurately verify a long list of entities in a single assessment imposes a high cognitive load, leading to inconsistent scores, while running it for every single entity is prohibitively expensive for RL. This leaves us in a predicament: simple methods are too inaccurate, and accurate methods are too impractical.

To overcome these intertwined challenges, we design a Hybrid Reward System. This system provides a nuanced, accurate, and cost-effective training signal, specifically tailored to the unique demands of our entity-intensive tasks while maintaining compatibility with standard benchmarks. It is composed of two core components: a granular, F-score-based reward for our synthesized tasks, and the retention of conventional reward functions for existing public benchmark data.

\paragraph{Granular F-Score for Entity-Intensive Tasks.}
For the approximately 500 entity-intensive QA pairs reserved for RL, we develop a fine-grained reward function based on the $\mathrm{ISR}$ metric. Recall that $\mathrm{ISR} = \frac{|R \cap O|}{|R|}$ (Equation~\ref{eq:ISR}) measures the recall of the retrieved entities.
Building upon $\mathrm{ISR}$ as our measure of recall, we designed a more comprehensive reward signal that also accounts for precision. An agent could otherwise achieve a high score by retrieving many irrelevant entities.

Furthermore, a practical reward function must gracefully handle minor semantic variations (e.g., ``USA'' vs. ``United States''). Therefore, we define soft versions of precision and recall by introducing a scoring function $s(e_o, e_r) \in [0, 1]$ that measures the semantic similarity between a retrieved entity $e_o \in O$ and a ground-truth entity $e_r \in R$. Instead of a monolithic judgment, we evaluate at the individual entity level. To balance accuracy and efficiency, we first categorize entities in the ground-truth set $R$ by their semantic type (e.g., person names, dates, organizations) and assign an appropriate evaluation modality $s$ to each category. For instance, person names might be evaluated using near-exact match to handle minor variations, while more abstract concepts might require a targeted LLM-as-a-Judge assessment.

Based on this semantic scoring function $s$, we define our soft recall $\gR_c$ (a generalization of $\mathrm{ISR}$) and soft precision $\gP$:
\begin{align}
    \gR_c &= \frac{1}{|R|} \sum_{e_r \in R} \max_{e_o \in O} s(e_o, e_r) \\
    \gP &= \frac{1}{|O|} \sum_{e_o \in O} \max_{e_r \in R} s(e_o, e_r)
\end{align}
This formulation credits the agent for finding entities that are semantically equivalent to the ground truth. We then aggregate $\gP$ and $\gR_c$ using a weighted F-score to compute the final reward $\mathcal{R}_{\w}$. This addresses potential biases in our synthesized ground-truth set $R$, which may be slightly over- or under-complete. The reward $\mathcal{R}_{\w}$ is defined as:
\begin{equation}
\mathcal{R}_{\w} = (1 + \omega^2) \frac{\gP \cdot \gR_c}{\omega^2 \gP + \gR_c}
\end{equation}
where $\omega$ is a hyperparameter that balances the importance of precision and recall. A value of $\omega > 1$ prioritizes recall (aligning more closely with the original goal of $\mathrm{ISR}$), while $\omega < 1$ emphasizes precision.

\paragraph{Hybrid Integration.}
For tasks originating from existing training QA, we retain their original, often binary, reward functions, which we denote as $\mathcal{R}_{\text{legacy}}$. Our final hybrid reward function, $\mathcal{R}_{\text{hybrid}}$, is therefore conditional on the task's origin, ensuring that the agent is evaluated appropriately for each data source:
\begin{equation}
\mathcal{R}_{\text{hybrid}}(\mathcal{H}_T, \mathcal{T}) =
\begin{cases}
\mathcal{R}_{\w}(O, R) & \text{if } \mathcal{T} \text{ is from \w} \\
\mathcal{R}_{\text{legacy}}(O, R) & \text{otherwise}
\end{cases}
\end{equation}
This hybrid reward signal provides rich, fine-grained feedback on our entity-intensive tasks while maintaining compatibility with established evaluation protocols. The agent's policy is then optimized against this comprehensive reward using Group Relative Policy Optimization (GRPO)~\citep{shao2024deepseekmathpushinglimitsmathematical}, enabling it to refine its information-seeking strategies effectively.

\begin{table}
\small
\centering
\caption{Results on multiple benchmarks.
All benchmarks except WideSearch report \texttt{Pass@1}.
WideSearch reports Success Rate (\texttt{SR}), \texttt{Row F1}, and \texttt{Item F1}.
\textbf{Bold} scores indicate the highest values among all open-source agents. \texttt{B} and \texttt{C} stand for base and comprehensive training setting.
}
\resizebox{\columnwidth}{!}{%
\setlength{\tabcolsep}{6pt} 
\renewcommand{\arraystretch}{1} 
\begin{tabular}{lccccccc}
\toprule
\multirow{2.5}{*}{\textbf{Model / Framework}} 
& \multirow{2.5}{*}{\textbf{BrowseComp}} 
& \multirow{2.5}{*}{\textbf{GAIA}} 
& \multirow{2.5}{*}{\textbf{xbench-DS}} 
& \multirow{2.5}{*}{\textbf{Seal-0}} 
& \multicolumn{3}{c}{\textbf{WideSearch}} \\
\cmidrule(lr){6-8}
 & & & & & \texttt{SR} & \texttt{Row F1} & \texttt{Item F1} \\
\midrule
\multicolumn{8}{c}{\cellcolor{blue!10} \textbf{\textit{Proprietary Agents}}} \\
\midrule
\texttt{Claude-4-Sonnet} & 12.2 & 68.3 & 64.6 & -- & 2.3 & 31.7 & 57.9 \\
\texttt{OpenAI-o3} & 49.7 & 70.5 & 66.7 & 18.9 & 4.5 & 34.0 & 52.6 \\
\texttt{OpenAI DeepResearch} & 51.5 & 67.4 & -- & -- & -- & -- & -- \\
\midrule
\multicolumn{8}{c}{\cellcolor{blue!10} \textbf{\textit{Open-Source Agents}}} \\
\midrule
\texttt{ASearcher-Web-32B} & 5.2 & 52.8 & 42.1 & -- & -- & -- & -- \\
\texttt{DeepDive-32B} & 14.8 & -- & 50.5 & -- & -- & -- & -- \\
\texttt{DeepDiver-V2-38B} & 13.4 & -- & 53.0 & -- & -- & -- & -- \\
\texttt{MiroThinker-32B-DPO-v0.2} & 13.0 & 64.1 & -- & -- & -- & -- & -- \\
\texttt{Kimi-K2-Instruct-1T} & 14.1 & 57.7 & 50.0 & -- & 1.1 & \textbf{29.7} & \textbf{54.4} \\
\texttt{WebExplorer-8B} & 15.7 & 50.0 & 53.7 & -- & -- & -- & -- \\
\texttt{WebDancer-QwQ-32B} & 3.8 & 51.5 & 38.3 & -- & 0.0 & 9.3 & 34.5 \\
\texttt{WebSailor-32B} & 10.5 & 53.2 & 53.3 & 21.3 & 0.0 & 2.1 & 5.5 \\
\texttt{WebShaper-QwQ-32B} & -- & 53.3 & 35.0 & -- & 0.0 & 9.9 & 31.5 \\
\midrule
\texttt{WebLeaper-Union B} & 22.1 & 69.9 & 62.3 & 35.1 & \textbf{4.0} & 22.2 & 34.5 \\
\texttt{WebLeaper-Reverse-Union B} & 23.0 & 67.0 & 66.0 & 37.2 & \textbf{4.0} & 25.8 & 40.8 \\
\texttt{WebLeaper-Reverse-Union C} & \textbf{38.8} & \textbf{73.2} & \textbf{72.0} & \textbf{48.6} & \textbf{4.0} & \textbf{31.0} & \textbf{48.8} \\
\bottomrule
\end{tabular}
}
\label{tab:main_result}
\vspace{-1em}
\end{table}

\paragraph{Policy Optimization with Hybrid Reward.}
The agent's policy, denoted $\pi_\theta$ parameterized by $\theta$, is optimized using GRPO. For each task $\mathcal{T}$ in our RL dataset, we sample a group of $k$ trajectories $\{\mathcal{H}_1, \dots, \mathcal{H}_k\}$ from the current policy $\pi_\theta$. Each trajectory $\mathcal{H}_i$ is assigned a reward $R_i = \mathcal{R}_{\text{hybrid}}(\mathcal{H}_i, \mathcal{T})$. Instead of using a learned value function, GRPO estimates the advantage for each trajectory by standardizing its reward relative to the others in the group:
\begin{equation}
\hat{A}_i = \frac{R_i - \text{mean}(\{R_j\}_{j=1}^k)}{\text{std}(\{R_j\}_{j=1}^k) + \epsilon_{\text{std}}}
\end{equation}
where $\epsilon_{\text{std}}$ is a small constant for numerical stability. This group-relative advantage $\hat{A}_i$ is applied to every timestep within the trajectory $\mathcal{H}_i$. The policy is then updated by minimizing a clipped surrogate objective, similar to PPO~\citep{schulman2017proximalpolicyoptimizationalgorithms}, which is averaged over all trajectories in the group and all timesteps in each trajectory. The GRPO loss function is:
\begin{equation}
\label{eq:grpo_loss}
\begin{aligned}
\mathcal{L}_{\text{GRPO}}(\theta) = -\mathbb{E}_{\{\mathcal{H}_i\}_{i=1}^k \sim \pi_{\theta}} \Bigg[ \frac{1}{k} \sum_{i=1}^{k} \frac{1}{|\mathcal{H}_i|} \sum_{t=1}^{|\mathcal{H}_i|} \Bigg( \min \bigg( r_{i,t}(\theta) \hat{A}_i,
\text{clip} \big( r_{i,t}(\theta), 1-\varepsilon, 1+\varepsilon \big) \hat{A}_i \bigg) \Bigg) \Bigg]
\end{aligned}
\end{equation}
where $r_{i,t}(\theta) = \frac{\pi_{\theta}(a_{i,t} \mid s_{i,t})}{\pi_{\text{old}}(a_{i,t} \mid s_{i,t})}$ is the importance sampling ratio at timestep $t$ of trajectory $i$, and $\varepsilon$ is the clipping hyperparameter. By optimizing this loss, the policy $\pi_\theta$ learns to favor actions that lead to trajectories with higher-than-average rewards within a sampled group, effectively internalizing the complex preferences defined by our $\mathcal{R}_{\text{hybrid}}$ function.


%% file: sections/4-experiments.tex
\section{Experiments}
\label{sec:Evaluation}

\subsection{Setup}
\label{sec:Setup}

\paragraph{Benchmarks} We conduct extensive evaluations of our method on five challenging QA benchmarks that demand complex information-seeking capabilities, namely BrowseComp~\citep{bc_en}, GAIA~\citep{mialon2023gaia}, xbench-DeepSearch (xbench-DS)~\citep{xbench}, Seal-0~\citep{pham2025sealqa}, and WideSearch~\citep{wong2025widesearch}. For GAIA, we adopt the 103-sample text-only validation subset~\citep{Li2025webthinker}, while for all other benchmarks, we utilize their complete test sets.


\begin{table}
\small
\centering
\caption{Ablation study on training results across different data sources (for efficiency considerations, we use the \texttt{WideSearch} (English subset) and \texttt{BrowseComp} (200 subset), while the full sets are used for the other benchmarks). Numbers in parentheses denote the difference compared to training only with the \texttt{WebSailor-V2-5k} data. $\dag$ denotes a mixed version that includes the \texttt{WebSailor-V2-5k} data.}
\label{tab:task_ablation}
\resizebox{\linewidth}{!}{
\begin{tabular}{lcccccc}
\toprule
{\textbf{Data Source}} & \textbf{BrowseComp} & \textbf{WideSearch} & \textbf{GAIA} & \textbf{Seal-0} & \textbf{xbench-DS} & \textbf{Avg.} \\
\midrule
\texttt{WebSailor-V2-5k} & 25.17 & 33.15 & 67.69 & 34.23 & 60.00 & 44.05 \\
\texttt{WebSailor-V2-10k} & 24.50 & 38.91 & 66.02 & 33.93 & 62.67 & 45.21 \\
\midrule
\texttt{Basic-5k}$^\dag$
  & 20.67 \,(\textcolor{red!70!black}{-4.50}) 
  & 32.26 \,(\textcolor{red!70!black}{-0.89}) 
  & 40.78 \,(\textcolor{red!70!black}{-26.91}) & 30.03 \,(\textcolor{red!70!black}{-4.20}) & 58.33 \,(\textcolor{red!70!black}{-1.67}) & 36.41 \,(\textcolor{red!70!black}{-7.64}) \\
\texttt{Union-5k}$^\dag$
  & 27.50 \,(\textcolor{green!60!black}{+2.33}) 
  & 41.70 \,(\textcolor{green!60!black}{+8.55}) 
  & 69.90 \,(\textcolor{green!60!black}{+2.21}) & 35.14 \,(\textcolor{green!60!black}{+0.82}) & 62.33 \,(\textcolor{green!60!black}{+2.33}) & 47.31 \,(\textcolor{green!60!black}{+3.26}) \\
\texttt{Reverse-Union-10k}$^\dag$
  & 27.67 \,(\textcolor{green!60!black}{+2.50}) & 44.07 \,(\textcolor{green!60!black}{+10.92}) & 66.99 (\textcolor{red!70!black}{-0.70}) & 37.24 \,(\textcolor{green!60!black}{+3.01}) & 66.00 \,(\textcolor{green!60!black}{+6.00}) & 48.39 \,(\textcolor{green!60!black}{+4.34}) \\
\bottomrule
\end{tabular}
}
\vspace{-1em}
\end{table}

\paragraph{Baselines} 
We select a representative set of mainstream and competitive information-seeking agents as our baselines, including proprietary agents (\texttt{Claude-4-Sonnet}~\citep{claude4}, \texttt{OpenAI-o3}~\citep{o3}, \texttt{OpenAI DeepResearch}~\citep{openaidr}) and open-source agents (\texttt{ASearcher}~\citep{asearcher}, \texttt{DeepDive}~\citep{lu2025deepdive}, \texttt{DeepDiver-V2}~\citep{deepdiver-v2}, \texttt{MiroThinker}~\citep{miromind2025mirothinker}, \texttt{Kimi-K2}~\citep{kimi-k2}, \texttt{WebExplorer}~\citep{liu2025webexplorer}, \texttt{WebDancer}~\citep{wu2025webdancerautonomousinformationseeking}, \texttt{WebSailor}~\citep{li2025websailornavigatingsuperhumanreasoning}, \texttt{WebShaper}~\citep{tao2025webshaper}).


\paragraph{Training Configurations} 
To maintain the basic deep search ability, we combine our data with 5,000 \texttt{WebSailor-V2}~\citep{li2025websailorv2bridgingchasmproprietary} data to train the model. We separately merge 5,000 \texttt{WebSailor-V2} data with \texttt{Basic}, \texttt{Union}, and \texttt{Reverse-Union} data of \w, which stimulates the IS ability to a larger degree  (with $\alpha$ in ISR set to 0.3 and $\beta$ in ISE set to 0.1). We employ \texttt{Qwen3-30B-A3B-Thinking-2507}\footnote{\url{https://huggingface.co/Qwen/Qwen3-30B-A3B-Thinking-2507}} as the base model, trained using the \texttt{Megatron} framework\footnote{\url{https://github.com/NVIDIA/Megatron-LM}}. This is our default base setting in which most experiments are conducted.

\paragraph{Comprehensive and Realistic Settings}

To more rigorously evaluate whether the training data of \w~can remain effective under more comprehensive and realistic scenarios, we introduce the comprehensive setting. We mix \w~ data into the corpus of \texttt{Tongyi-DeepResearch-30B-A3B}, covering both the supervised fine-tuning and reinforcement learning stages, to examine its overall impact on performance. 
It is worth noting that this serves only as a supplementary setting applied in certain experimental sections. Unless otherwise specified, we adopt the base \w~ experimental configuration by default.


\paragraph{Evaluation Metrics and Inference Hyper-parameters} The overall evaluation follows the settings specified by each benchmark. For BrowseComp, GAIA, xbench-DS, and Seal-0, we report the \texttt{pass@1} scores obtained via LLM-as-a-judge evaluation as the final results. For WideSearch, we report the success rate (\texttt{SR}) for fully retrieving all target results, along with two F1 scores—\texttt{Row F1} and \texttt{Item F1}—which are computed using a combination of string matching and LLM-as-a-judge evaluation, in alignment with the official evaluation protocol. During LLM inference, we configure the sampling parameters (temperature and top‑\textit{p}) to 0.6 and 0.95, respectively.


\subsection{Overall Performance}
\label{sec:main_results}

\noindent\textbf{Base Setting}
As shown in Table~\ref{tab:main_result}, \texttt{WebLeaper} achieves state-of-the-art performance compared to mainstream open-source agents on five challenging information-seeking QA benchmarks. Notably, on benchmarks other than BrowseComp and WideSearch, it even delivers performance comparable to, or surpassing, that of agents built on \texttt{Claude-4-Sonnet} and \texttt{OpenAI-o3}. Even on the highly challenging BrowseComp benchmark, \texttt{WebLeaper} significantly outperforms \texttt{Kimi-K2-Instruct-1T}, despite the latter having a much larger parameter scale. It is also worth noting that the \texttt{Reverse-Union} data, which incorporates greater task complexity on top of the \texttt{Union} data, employs an fuzz strategy that further facilitates the model’s ability to integrate information-seeking with planning and reasoning, thereby enhancing its overall information-seeking QA capability.

Overall, the observed performance improvements validate that our proposed approaches—entity-intensive task synthesis and information-guided trajectory construction—significantly enhance the agent’s information-seeking capabilities, even under a modest parameter budget.

\noindent\textbf{Comprehensive Setting}
We also train our method on the comprehensive setting, and compare it to more competitive methods. The results are shown in Figure~\ref{fig:main_results}. 
\w~ reaches 73.2 on GAIA, 38.8 on BrowseComp, and 72.0 on xbench-DeepResearch. On the harder WideSearch benchmark, \w~ also attains the highest Success Rate and Item-F1, clearly outperforming all competitors. These results demonstrate that our approach generalizes well and remains effective even when evaluated under the comprehensive and realistic training setting.

\begin{figure}[t]
    \centering
    \includegraphics[width=1\textwidth]{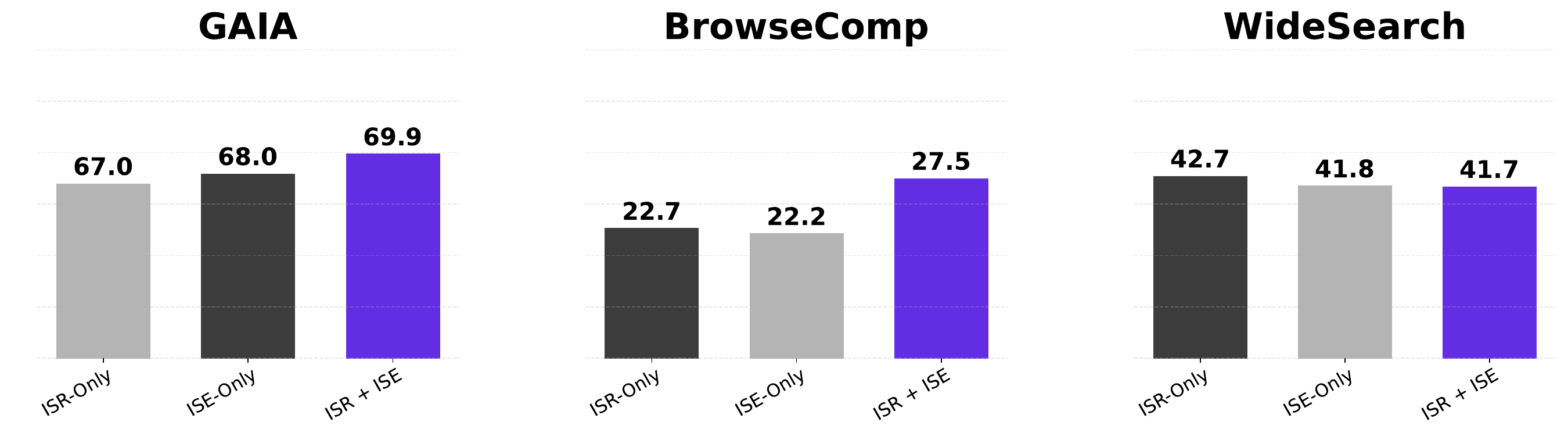}
    \caption{Ablation study results on information-guided trajectory construction strategies.}
    \label{fig:traj_ablation_bar_plot}
    \vspace{-1em}
\end{figure}

\subsection{Capability Gains Induced by Entity-Intensive Task Synthesis}

To investigate the effectiveness of our entity-intensive task synthesis method, we conduct a comparative analysis against training solely on the \texttt{WebSailor-V2} dataset (using 5,000 and 1,000 samples, respectively), a synthetic corpus specifically designed to stimulate the agent’s deep search capability.

As shown in Table~\ref{tab:task_ablation}, we investigate the impact of different entity-intensive task synthesis strategies through an ablation study on all these benchmarks. The \texttt{Basic} setting exhibits substantial drops across all three datasets compared to \texttt{WebSailor-V2-5k}. This poor performance can be attributed to the inherent limitations of the Basic data construction method: tasks generated under this setting tend to be overly simple, allowing the model to infer complete answers from only a few information sources. Such shortcut patterns encourage the model to overfit to superficial cues rather than learning to integrate diverse information, ultimately impairing generalization.

In contrast, the \texttt{Union} strategy consistently outperforms \texttt{WebSailor-V2-5k}, achieving an average improvement of $+3.26$. By combining heterogeneous information sources and increasing the complexity of task construction, \texttt{Union} mitigates the shortcut problem inherent in \texttt{Basic}, forcing the model to reason over dispersed and complementary evidence. This leads to more robust performance across datasets and demonstrates the effectiveness of the proposed data construction approach.

Furthermore, compared to \texttt{Union}, \texttt{Reverse-Union} introduces a certain degree of reasoning complexity into the information-seeking process, making it more challenging for the model to readily identify where to begin entity retrieval. This design particularly enhances the model’s planning and decision-making capabilities in information-seeking tasks. The improvement in these abilities is clearly reflected in performance, leading to substantial and widespread gains across all benchmarks.

\subsection{Impact of Information-Guided Trajectory Construction}

We compare the proposed information-guided trajectory construction strategies across \texttt{ISR-Only}, \texttt{ISE-Only}, and \texttt{ISR+ISE} on three representative benchmarks—GAIA, BrowseComp, and WideSearch—to examine the independent and combined effects of ISE and ISR.

On GAIA and BrowseComp, \texttt{ISR+ISE} achieves the best performance, suggesting that integrating precision and efficiency constraints produces trajectories that are both goal-directed and concise, thereby reducing redundant exploration. This indicates that in more complex browsing tasks, relevance and efficiency constraints complement each other to generate higher-quality trajectories.

In contrast, on WideSearch, the three strategies deliver comparable results, with performance differences falling within the margin of variance. This suggests that for broad search tasks, the specific choice of trajectory filtering plays a less critical role—likely because training on entity-intensive synthesized data already provides strong broad search capabilities.

\subsection{Joint Gains in Efficiency and Effectiveness}

\begin{figure}[t]
    \centering
    \includegraphics[width=0.9\textwidth]{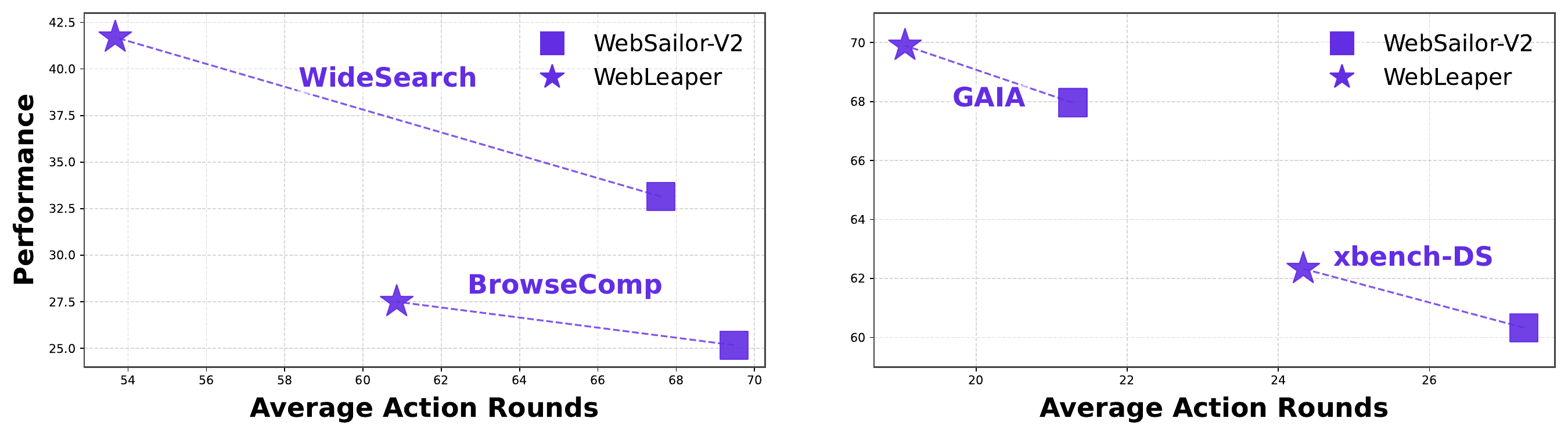}
    \caption{Effectiveness and efficiency comparison between \w{} and \texttt{WebSailor-V2}.}
    \label{fig:tool_call_performance_scatter}
    \vspace{-1em}
\end{figure}

As illustrated in Figure~\ref{fig:tool_call_performance_scatter}, \w{} consistently outperforms the baseline in terms of both effectiveness and efficiency. In the WideSearch and BrowseComp benchmarks, our approach achieves markedly higher performance scores while requiring fewer average action rounds, indicating that the search process is not only more accurate but also more efficient. Similarly, in the GAIA and xbench-DS tasks, our method improves effectiveness while simultaneously reducing the operational cost. This demonstrates that our design enables a more targeted search strategy, resulting in reduced interaction steps without sacrificing—and in fact enhancing—the quality of the results. 

Overall, these results validate that our proposed method achieves superior joint optimization of information-seeking efficiency and task performance compared to the baseline. This reflects our key insight: an agent should not merely learn to search, but rather learn to search efficiently and wisely, thereby achieving a better balance between efficiency and effectiveness.

\subsection{Reinforcement Learning using \w{}}

\begin{table*}[h!]
\tiny
\centering
\caption{RL Results on comprehensive setting.
All benchmarks except WideSearch report Avg \texttt{Pass@1} from 3 rollouts.
WideSearch reports Success Rate (\texttt{SR}), \texttt{Row F1}, and \texttt{Item F1}.
}
\resizebox{\columnwidth}{!}{%
\setlength{\tabcolsep}{6pt} 
\renewcommand{\arraystretch}{1.1} 
\begin{tabular}{lcccccc}
\toprule
\multirow{2.5}{*}{\textbf{}} 
& \multirow{2.5}{*}{\textbf{BrowseComp}} 
& \multirow{2.5}{*}{\textbf{GAIA}} 
& \multirow{2.5}{*}{\textbf{xbench-DS}} 
& \multicolumn{3}{c}{\textbf{WideSearch}} \\
\cmidrule(lr){5-7}
 & & & & \texttt{SR} & \texttt{Row F1} & \texttt{Item F1} \\
\midrule
\texttt{SFT} & 37.80 & 69.9 & 69.0 & 1.5 & 23.0 & 45.4 \\
\midrule
\texttt{SFT+RL} & 38.8 \,(\textcolor{green!60!black}{+1.0}) & 73.2 \,(\textcolor{green!60!black}{+3.3}) & 72.0 \,(\textcolor{green!60!black}{+3.0}) & 4.0 \,(\textcolor{green!60!black}{+2.5}) & 31.0 \,(\textcolor{green!60!black}{+8.0}) & 48.5 \,(\textcolor{green!60!black}{+3.1}) \\
\bottomrule
\end{tabular}%
}
\label{tab:rl}
\end{table*}

\begin{figure}[t]
    \centering
    \includegraphics[width=0.7\textwidth]{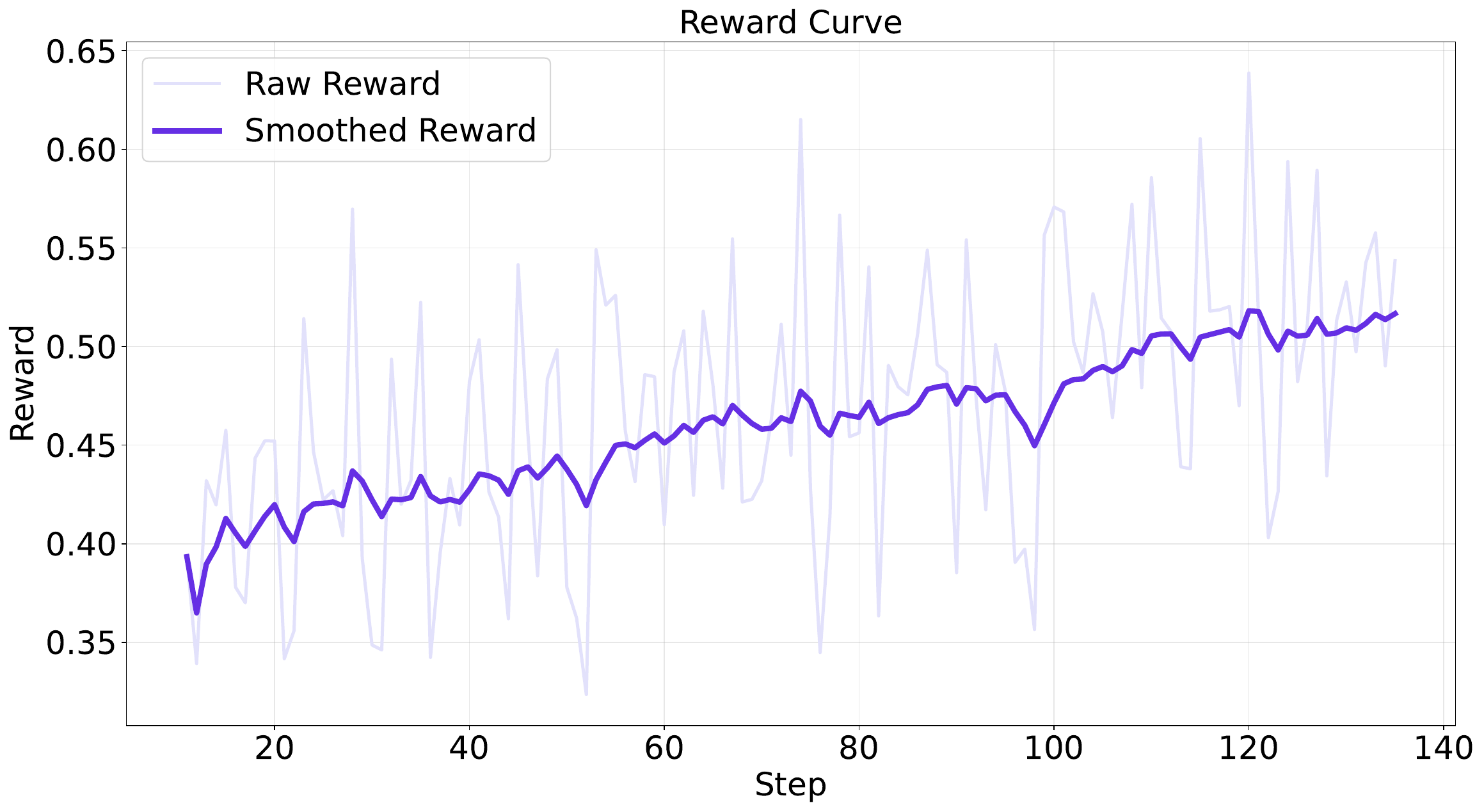}
    \caption{Figure shows the training curve of the hybrid reward system, indicating that using the \w{} data leads to a stable increase in reward. We terminated the experiment at 135 steps when web access resources were exhausted and evaluated the results at this point.}
    \label{fig:reward_curve}
    \vspace{-1em}
\end{figure}

We further evaluate our approach through reinforcement learning, adopting the \textit{Additional Settings for More Comprehensive and Realistic Training} (see Section~\ref{sec:Setup}) where \w{} data is mixed into a larger training corpus for both SFT and subsequent RL stages. As demonstrated by the results in Table~\ref{tab:rl} and Figure~\ref{fig:reward_curve}, using \w{} data for RL yields consistent and significant improvements. The results table shows that after RL fine-tuning, the model comprehensively surpasses the SFT-only baseline across all benchmarks. 

This positive performance trend is echoed by the reward curve in Figure~\ref{fig:reward_curve}, which exhibits a stable and continuous upward trajectory throughout the training process. This indicates that the model is effectively learning from the reward signals derived from the \w{} data, progressively refining its information-seeking strategy towards greater efficiency and accuracy. Even with the experiment concluding at 135 steps, the clear learning trend underscores the potential for further gains.

The results strongly validate the effectiveness of the \w{} dataset. It not only serves as a robust foundation for supervised fine-tuning but also provides a high-quality signal for RL, successfully guiding the agent to master more sophisticated and optimal information-seeking behaviors.

%% file: sections/5-related_work.tex
\section{Related Work}
\label{sec:related_work}

\subsection{Information Seeking Agent}
\label{sec:InformationSeekingAgent}

LLM-powered information-seeking agents can be broadly categorized into 3 streams: (1) enhancing core models via supervised fine-tuning~\citep{zeng2023agenttuning,wu2025webdancerautonomousinformationseeking,li2025websailornavigatingsuperhumanreasoning,li2025websailorv2bridgingchasmproprietary,tao2025webshaper,su2025scalingagentscontinualpretraining,fang2025generalagenticintelligenceenvironment}; (2) advancing agent architecture for improved planning and robustness~\citep{qiao2025webresearcher,xu2025amemagenticmemoryllm,li2025raspberry}; and (3) developing multi-agent systems for collaborative problem-solving~\citep{wu2023autogen,hong2023metagpt}.
Our work aligns with the first category but addresses a key limitation. Prior methods often train on tasks focused on correctness with single-fact answers, which is insufficient for large-scale information gathering. We posit that the number of entities in an answer—its entity richness—is a critical dimension for evaluating an agent's completeness and efficiency. This paper aims to bridge this gap by creating and utilizing entity-rich QA data to enhance agent capabilities for comprehensive information acquisition.

\subsection{Agent Data Synthesis}
\label{sec:AgentDataSynthesis}

Synthetic data generation is pivotal for agent training, with primary applications in tool use~\citep{wu2025webdancerautonomousinformationseeking, tao2025webshaper, shen2025ragsynthsyntheticdatarobust,fang2025towards}, code generation~\citep{jimenez2024swebench,shen2025shortcutsbench,xu2025kodcode, shao2025case2code}, and GUI automation~\citep{xu2025agenttrek, sun-etal-2025-os, pahuja2025explorer}. These efforts primarily combat data scarcity.
Within the information-seeking domain, existing data synthesis approaches increase task difficulty through multi-step reasoning~\citep{wu-etal-2025-webwalker,wu2025webdancerautonomousinformationseeking,tao2025webshaper} or long-horizon planning~\citep{qiao2025webresearcher}. We contend that such methods often overlook the semantic richness of the training data itself. In contrast, our approach centers on synthesizing QA data with high entity-level complexity. We hypothesize that this focus on data semantics is a crucial and complementary path to improving agent reasoning and world knowledge alignment.

%% file: sections/6-conclusion.tex
\section{Conclusion}
\label{sec:conclusion}

In this paper, we addressed the critical challenge of low search efficiency in LLM-based information-seeking agents, a bottleneck that constrains their overall performance. We argued that the sparsity of target entities in conventional training tasks is a primary contributor to this inefficiency. To overcome this, we introduced \w, a novel framework for constructing entity-intensive IS tasks and generating efficient solution trajectories. By formulating IS as a tree-structured reasoning problem and systematically increasing task complexity through our \texttt{Basic}, \texttt{Union}, and \texttt{Reverse-Union} task synthesis variants, we created a rich training environment. Furthermore, our information-guided trajectory curation, using ISR and ISE metrics, ensures that the agent learns from solutions that are both accurate and efficient. Our extensive experiments demonstrated that \w consistently improves performance across five challenging benchmarks, validating that enhancing search efficiency is a powerful lever for boosting the overall capabilities of IS agents.

%% file: sections/appendix.tex
\newpage
\section{Appendix}
\label{sec:appendix}

\subsection{Declaration on the Use of LLMs}
We declare that the use of LLMs during the preparation of this manuscript was strictly limited to language-related assistance, such as sentence refinement and grammatical correction. All substantive content was independently authored by the authors and rigorously reviewed and verified following any LLM-assisted modifications. During the experiments, all usage of LLMs was solely for academic research purposes, with no inappropriate applications. Detailed experimental settings are provided in the Experiments section of this paper. No other reliance on LLMs is involved in this work.

\subsection{Proof of Proposition~\ref{prop:var_reduction}}
\label{sec:appendix_proof_ise}

This appendix provides the detailed mathematical derivation for Proposition~\ref{prop:var_reduction}, as presented in Section~\ref{subsec:quantifying_collection}. The purpose of this proof is to formally establish that the variance of the Information-Seeking Efficiency (ISE) metric is inversely proportional to $n$, the number of required entities. This property, $\mathrm{Var}(\mathrm{ISE}) = \mathcal{O}(1/n)$, demonstrates that ISE becomes an increasingly stable and reliable performance measure as the complexity of the task (i.e., the size of $n$) grows.

\begin{proof}
Let $X_i$ be the number of steps the agent takes to discover the $i$-th new entity in the required set $R$. We assume $\{X_i\}_{i=1}^n$ are independent and identically distributed (i.i.d.) random variables with mean $\mathbb{E}[X_i] = \mu$ and variance $\mathrm{Var}(X_i) = \sigma^2$.

The total number of steps is $T = \sum_{i=1}^n X_i$. Let $\overline{X}$ be the average number of steps to find one required entity, defined as $\overline{X} = \frac{1}{n} \sum_{i=1}^n X_i = \frac{T}{n}$. By definition, $\mathrm{ISE} = n/T = 1/\overline{X}$.

From the properties of i.i.d. random variables, the mean and variance of $\overline{X}$ are:
\begin{align}
\mathbb{E}[\overline{X}] &= \mu, \\
\mathrm{Var}(\overline{X}) &= \frac{\sigma^2}{n}. \label{eq:var_xbar_appendix}
\end{align}
We are interested in the variance of $\mathrm{ISE}$, which is a function of the random variable $\overline{X}$. Let this function be $f(\overline{X}) = 1/\overline{X}$. We can approximate the variance of $\mathrm{ISE}$ using the Delta method, which states that for a function $f$ with a non-zero derivative at $\mu$:
\[
\mathrm{Var}(f(\overline{X})) \approx \left(f'(\mathbb{E}[\overline{X}])\right)^2 \mathrm{Var}(\overline{X}).
\]
First, we compute the derivative of $f(x) = 1/x$, which is $f'(x) = -x^{-2}$. Evaluating this derivative at the mean $\mu$:
\[
f'(\mu) = -\mu^{-2}.
\]
Now, substituting this and the variance of $\overline{X}$ from Equation~(\ref{eq:var_xbar_appendix}) into the Delta method formula:
\[
\mathrm{Var}(\mathrm{ISE}) \approx \left( -\mu^{-2} \right)^{2} \cdot \frac{\sigma^{2}}{n}
    = \frac{1}{\mu^{4}} \cdot \frac{\sigma^{2}}{n}
    = \mathcal{O}\left( \frac{1}{n} \right).
\]

This completes the proof.
\end{proof}

\subsection{Data Statistics}
\label{sec:data_statistics}

\begin{figure}[ht]
\centering
\includegraphics[width=0.6\linewidth]{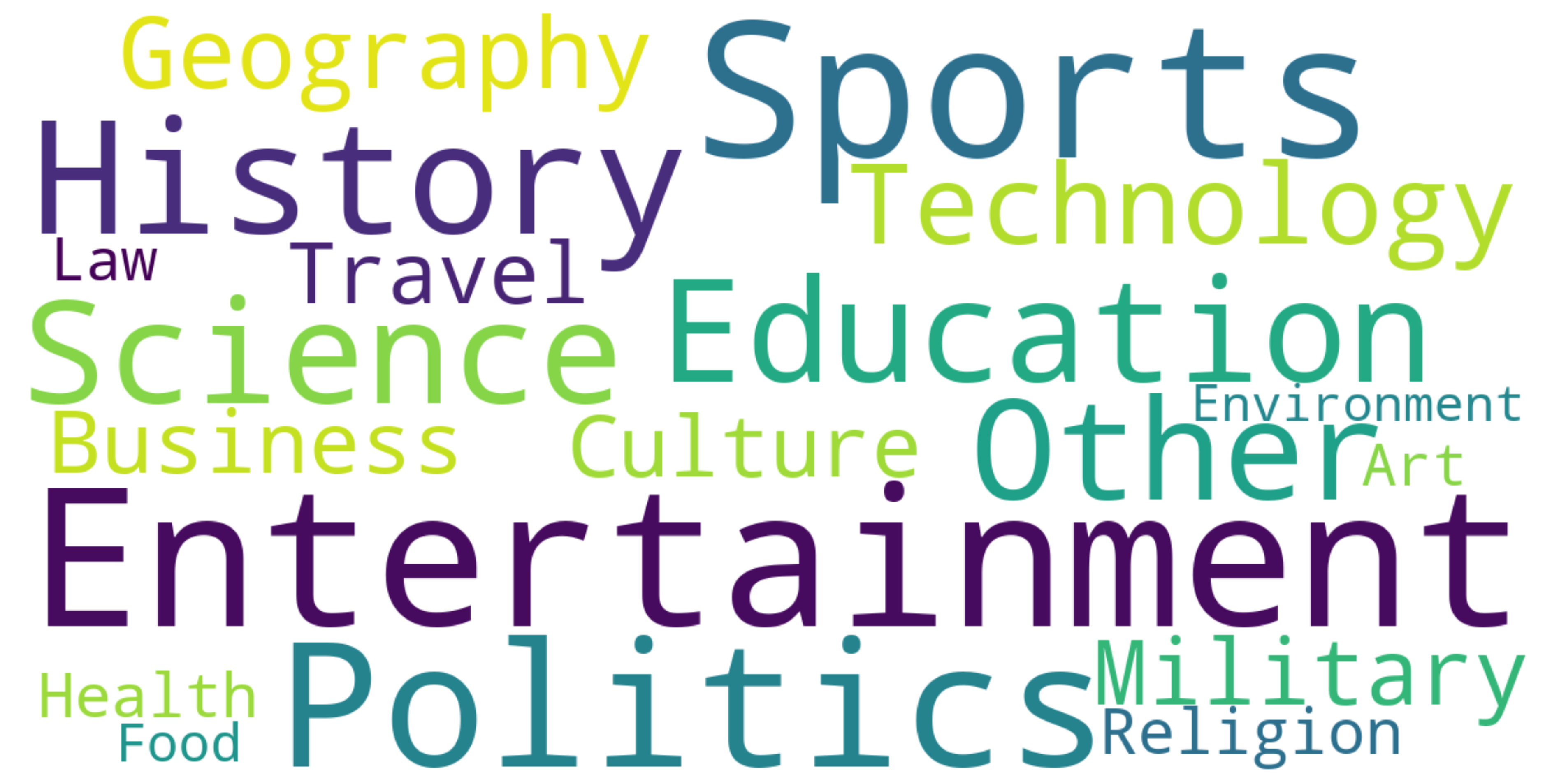}
\caption{The distribution of our training data.}
\label{fig:domain_distribution}
\vspace{-1em}
\end{figure}

Figure~\ref{fig:domain_distribution} illustrates the distribution of our training data.

\begin{figure}
\centering
\includegraphics[width=0.8\linewidth]{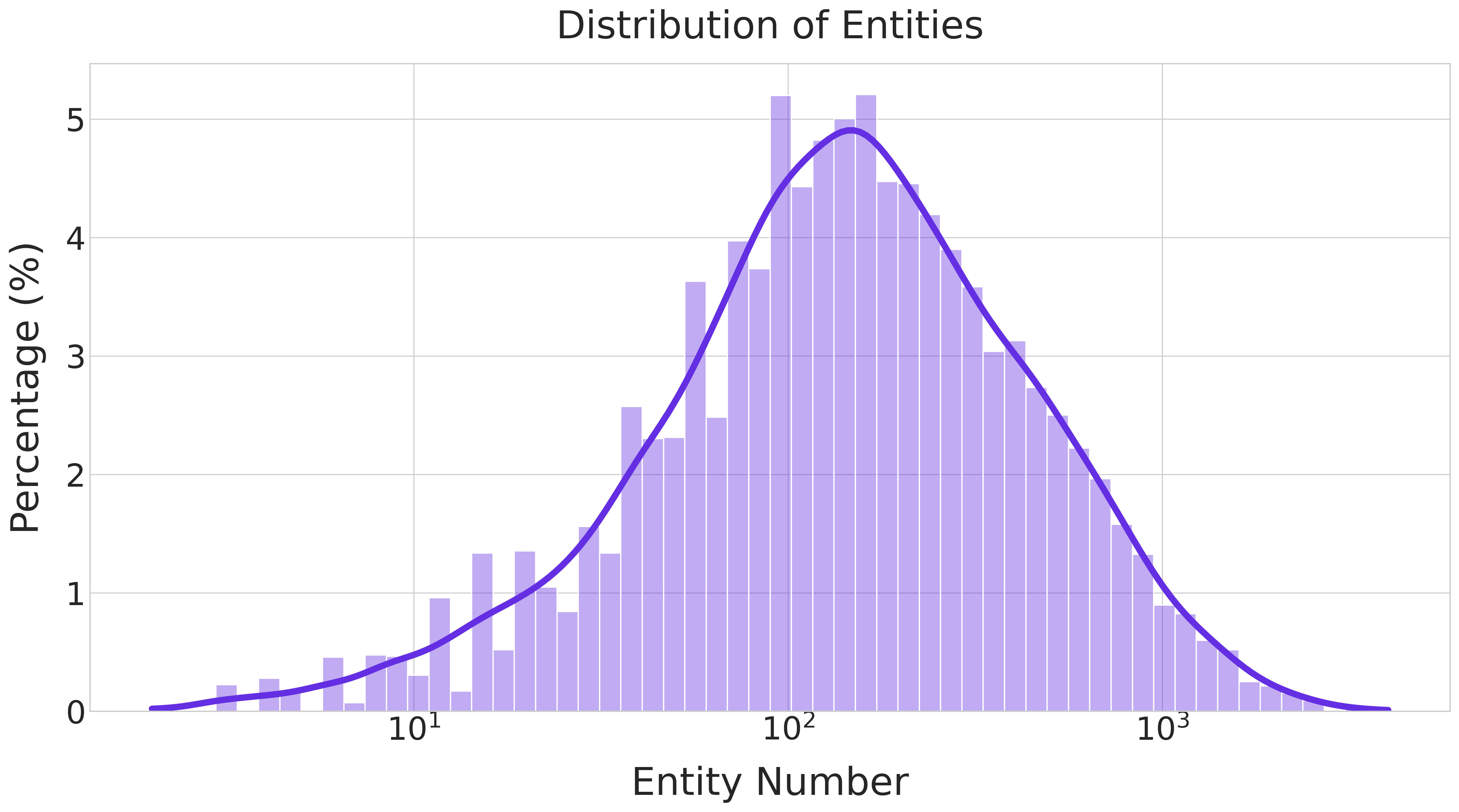}
\caption{Entity Count Distribution in Training Data. A significant portion of our samples contains at least 100 entities, underscoring the inherent difficulty of our dataset. This complexity, as formalized in Equation~\eqref{eq: V}, is crucial for robustly measuring efficiency, which in turn contributes to improved overall performance.}
\label{fig:entity_distribution}
\vspace{-1em}
\end{figure}

Figure \ref{fig:entity_distribution} displays the entity count distribution of our training data. A significant portion of our samples contain at least 100 entities, underscoring the inherent difficulty of our dataset. As formalized in Equation \ref{eq: V}, this complexity is crucial for robustly measuring efficiency, which in turn leads to improved overall performance.

\subsection{Data Cleaning and Basic Task Construction}
\label{sec:appendix_cleaning_basic}

This section elaborates on the data processing and construction methodology for the \texttt{Basic} version tasks introduced in Section~\ref{sec:base_dataset_construction}.

\paragraph{Rationale for Tree Structure}
In information-seeking tasks, the reasoning structure is paramount. We chose a tree structure for our basic tasks because it offers a compact and hierarchical organization of entities. This structure is highly efficient for representing a large number of interconnected entities that stem from a common query concept, mirroring many real-world information-gathering scenarios. A reasoning tree is composed of a root (question entity) and a set of subtrees, where each subtree represents a cohesive unit of information.

\paragraph{Multi-Stage Table Cleaning}
To ensure the quality and suitability of the data used for task synthesis, we crawled approximately 2 million tables from Wikipedia and subjected them to a rigorous multi-stage cleaning procedure. This was essential because raw web tables are often noisy and inconsistent. The stages were as follows:

\begin{itemize}[left=0.2cm]
\item \textbf{Size Filtering:} We first discarded tables that were either too small (fewer than 10 rows or 3 columns) to capture meaningful relational information, or too large (more than 200 rows or 20 columns) to be processed efficiently and form a coherent task.
\item \textbf{Semantic and Structural Filtering:} We then removed semantically irrelevant columns that frequently appear in web tables, such as those containing serial numbers, notes, or references. Tables with significant formatting errors (e.g., numerous merged cells that disrupt the relational structure) were also excluded.
\item \textbf{Isomorphism and Homogeneity:} Finally, we retained only groups of isomorphic tables (tables sharing the same column headers and structure). This step was crucial for ensuring structural homogeneity across our dataset, which is a prerequisite for identifying common subtree structures needed for the \texttt{Union} operation described later.
\end{itemize}

The resulting collection contains clean, well-structured tables with a set of meaningful fields as columns and multiple rows, where each row can be transformed into a subtree.

\paragraph{Reasoning Tree Population}
To construct the three-layer reasoning tree from a single table, we populate the layers as follows:
\begin{itemize}[left=0.4cm]
\item \textbf{First Layer (Question Entities):} Entities mentioned in the table's title or caption are extracted to form the root of the tree.
\item \textbf{Second Layer (Roots of Subtrees):} We employ an LLM to analyze the table's columns and select one that contains no duplicate entries. This column is treated as the key, and its values become the second-layer entities of the tree. Each of these entities serves as the root of a subtree. The LLM is effective at identifying columns like `Name' or `Title' that serve this unique identification purpose.
\item \textbf{Third Layer (Leaves of Subtrees):} The values in the remaining columns of the table constitute the third layer, representing the leaf entities associated with each second-layer entity.
\end{itemize}

\subsection{Maximal Union Algorithm for Task Synthesis}
\label{sec:appendix_maximal_fusion}

This section provides the formal definition and algorithmic implementation for discovering maximal union groups, as introduced in Section~\ref{sec:union_synthesis}. The core of our approach is to reformulate the search for compatible reasoning trees as a Maximal Biclique Enumeration (refer to ~\ref{alg:appendix_fusion} problem on a bipartite graph.

\textbf{Problem Formulation}

Let $\mathcal{T}_{\text{base}} = \{T_1, T_2, \dots, T_N\}$ be our collection of basic reasoning trees. We first construct a bipartite graph $G=(U, V, E)$, where $U = \mathcal{T}_{\text{base}}$ is the set of all trees, and $V$ is the set of all unique relation names found within the subtrees across all trees in $\mathcal{T}_{\text{base}}$. An edge $(T_i, v_j) \in E$ exists if the relation $v_j$ is present in any subtree of tree $T_i$ (i.e., $v_j \in \text{Rel}(T_i)$, where $\text{Rel}(T_i) = \bigcup_k \text{Rel}(S_{i,k})$).

In this construction, a \textit{maximal union} directly corresponds to a \textit{maximal biclique} $(\mathcal{U}, \mathcal{V})$, where $\mathcal{U} \subseteq U$ is a set of trees and $\mathcal{V} \subseteq V$ is a set of their common relations. Our goal is to find all such maximal bicliques that satisfy certain size and semantic constraints. Formally, we seek to find all maximal pairs $(\mathcal{U}, \mathcal{V})$ that satisfy:
\begin{equation}
\label{eq:optimization}
\begin{aligned}
\text{find maximal} \quad & (\mathcal{U}, \mathcal{V}) \\
\text{subject to} \quad & \forall T_i \in \mathcal{U}, \mathcal{V} \subseteq \text{Rel}(T_i), \\
& |\mathcal{U}| \ge k_{\min}, |\mathcal{V}| \ge m_{\min}.
\end{aligned}
\end{equation}

Here, maximality means that no other tree can be added to $\mathcal{U}$ and no other relation can be added to $\mathcal{V}$ without violating the biclique property. Solving this by reformulating it as a standard maximal biclique enumeration problem is computationally efficient compared to an exhaustive search.

\textbf{Algorithm and Implementation Details}

\begin{itemize}[left=0.4cm]
    \item \textbf{Input:} A collection of base reasoning trees $\mathcal{T}_{\text{base}}$; a minimum number of trees for a valid union, $k_{\min}$; a minimum number of common relations, $m_{\min}$.
    \item \textbf{Goal:} To find all \textit{maximal union groups}, which are the solutions $(\mathcal{U}, \mathcal{V})$ to Eq.~(\ref{eq:optimization}) that also satisfy the semantic matching criteria below.
    \item \textbf{Subtree Relation Matching Criteria:} To ensure the semantic coherence of unions, we impose strict matching criteria. For relations connecting the second and third layers, we require they share the same standardized name, data type, and domain. For the second-layer entities themselves (the roots of the subtrees), we relax this constraint, requiring only a match in data type and domain. This flexibility allows for the union of trees with conceptually similar but differently named second-layer entities (e.g., fusing a tree where entities are 'Authors' with another where they are 'Writers').
    \item \textbf{Output:} A set of maximal union groups $\mathcal{F}$, where each element is a tuple $\langle U', V' \rangle$ that meets the specified criteria.
\end{itemize}

The process is detailed in Algorithm~\ref{alg:appendix_fusion}.

\begin{algorithm}[ht!]
\caption{Maximal Union Identification Algorithm}
\label{alg:appendix_fusion}
\KwIn{A collection of base reasoning trees $\mathcal{T}_{\text{base}}$, minimum trees $k_{\min}$, minimum common relations $m_{\min}$.}
\KwOut{A set of maximal union groups $\mathcal{F}$.}

$\mathcal{F} \leftarrow \emptyset$\;

\tcp{1. Construct the bipartite graph from trees and subtree relations}
Let $U$ be the set of trees from $\mathcal{T}_{\text{base}}$ and $V$ be the set of unique standardized relation names found within the subtrees of all trees in $\mathcal{T}_{\text{base}}$\;
Construct the graph $G=(U, V, E)$ where an edge $(u, v) \in E$ exists if tree $u$ contains the relation $v$ in its subtrees (i.e., $v \in \text{Rel}(u)$)\;

\tcp{2. Enumerate maximal bicliques from the graph}
$\mathcal{B} \leftarrow \text{EnumerateMaximalBicliques}(G)$\;
\tcp*{Leverages standard algorithms like MICA or Eclat}

\tcp{3. Filter and validate bicliques to form final union groups}
\For{each maximal biclique $(U', V')$ in $\mathcal{B}$}{
    \tcp{Check size constraints from Eq. (1)}
    \If{$|U'| < k_{\min}$ or $|V'| < m_{\min}$}{
        \textbf{continue}\;
    }

    \tcp{Validate semantic compatibility of second-layer entities}
    Let $T_{\text{id}}, D_{\text{id}}$ be the type and domain of the second-layer entities of the first tree in $U'$\;
    is\_compatible $\leftarrow$ \textbf{true}\;
    \For{each tree $u \in U'$}{
        \If{$u$'s second-layer entity type $\neq T_{\text{id}}$ or domain $\neq D_{\text{id}}$}{
            is\_compatible $\leftarrow$ \textbf{false}\;
            \textbf{break}\;
        }
    }

    \tcp{If all checks pass, add to the set of valid union groups}
    \If{is\_compatible}{
        $\mathcal{F} \leftarrow \mathcal{F} \cup \{\langle U', V' \rangle\}$\;
    }
}
\KwRet{$\mathcal{F}$}\;
\end{algorithm}

\subsection{Detailed Examples of Task Synthesis}
\label{sec:appendix_examples}

This section provides detailed explanations and reasoning walkthroughs for the examples of the three task synthesis versions presented in Section~\ref{sec:method} and Figure~\ref{fig:overview}.

\subsubsection{\textbf{Version-I: Basic}}
\label{sec:example Version-I: Basic}

The goal of the basic version is to create a task with a clear, hierarchical reasoning structure derived from a single, self-contained set of entities.

\textbf{Example Question:} \textit{Who were the Nobel Prize winners in Literature between 1980 and 1990? Please include their name, country, award year, and gender.}

\textbf{Construction Process:}
The task is constructed from a single Wikipedia table, forming a reasoning tree. The layers shown in Figure~\ref{fig:overview}(a) are populated as follows:
\begin{itemize}[left=0.4cm]
\item \textbf{First Layer (question entities):} Derived from the table's title and a specified constraint, forming the query's scope: {\emph{Literature Nobel Prize, year 1980–1990}}.
\item \textbf{Second Layer (subtree roots):} Populated from the table's key column (e.g., author names): {\emph{Czesław Miłosz, William Golding}, \ldots}.
\item \textbf{Third Layer (subtree leaves):} Consists of values from the remaining columns, representing attributes for each second-layer entity. For example: {\emph{man, Poland, 1980}} for Czesław Miłosz. The edges connecting the second to the third layer represent relations like `has\_gender', `has\_country', `has\_award\_year'.
\end{itemize}

\textbf{Reasoning Path:}
An agent is expected to follow this hierarchical structure:
\begin{itemize}[left=0.4cm]
\item \textbf{Identify Scope:} Recognize the ``Question Entities'' from the query: {\emph{Nobel Prize in Literature, 1980–1990}}.
\item \textbf{Retrieve Second-Layer Entities:} Retrieve the second-layer entities, which are the authors: {Czesław Miłosz, William Golding, ...}.
\item \textbf{Gather Attributes:} For each second-layer entity, follow the relations to retrieve their associated third-layer entities, such as {Poland, 1980, man} for Czesław Miłosz.
\end{itemize}

\subsubsection{\textbf{Version-II: Union}}
\label{sec:example Version-II: Union}

This version increases structural complexity by requiring the agent to perform relational operations across distinct reasoning trees.

\textbf{Example Question:} \textit{Which authors have won both the Nobel Prize in Literature and the Booker Prize? For each, provide their name, nationality and the year they won the Nobel.}

\textbf{Construction Process:}
Once a maximal union is identified (e.g., between the reasoning trees for ``Nobel Prize laureates'' and ``Booker Prize winners,'' which share common relations like ``has\_nationality'' within their subtrees), an LLM generates a task requiring information integration. The LLM is prompted to find an interesting relationship, such as the intersection of the two sets of second-layer entities (authors), and then weave this logic into a natural language question.

\textbf{Reasoning Path:}
The task is constructed from a \textit{maximal union} of two distinct reasoning trees. To solve this, an agent must:
\begin{itemize}[left=0.4cm]
    \item \textbf{Retrieve First Entity Set:} Identify the first concept, ``Nobel Prize in Literature,'' and retrieve the full set of corresponding second-layer entities from the first tree, $R_{\text{Nobel (T1)}}$.
    \item \textbf{Retrieve Second Entity Set:} Identify the second concept, ``Booker Prize,'' and retrieve its full set of second-layer entities from the second tree, $R_{\text{Booker (T2)}}$.
    \item \textbf{Find Intersection:} Perform a relational join to find the intersection of the two sets of second-layer entities based on name. The final ``Target Entities'' are the entities present in both sets, such as \{\textit{William Golding}, \textit{J.M. Coetzee}, \dots\}, along with their requested third-layer attributes.
\end{itemize}

\subsubsection{\textbf{Version-III: Reverse-Union}}
\label{sec:example Version-III: Reverse-Union}

\begin{figure}
\centering
\includegraphics[width=\linewidth]{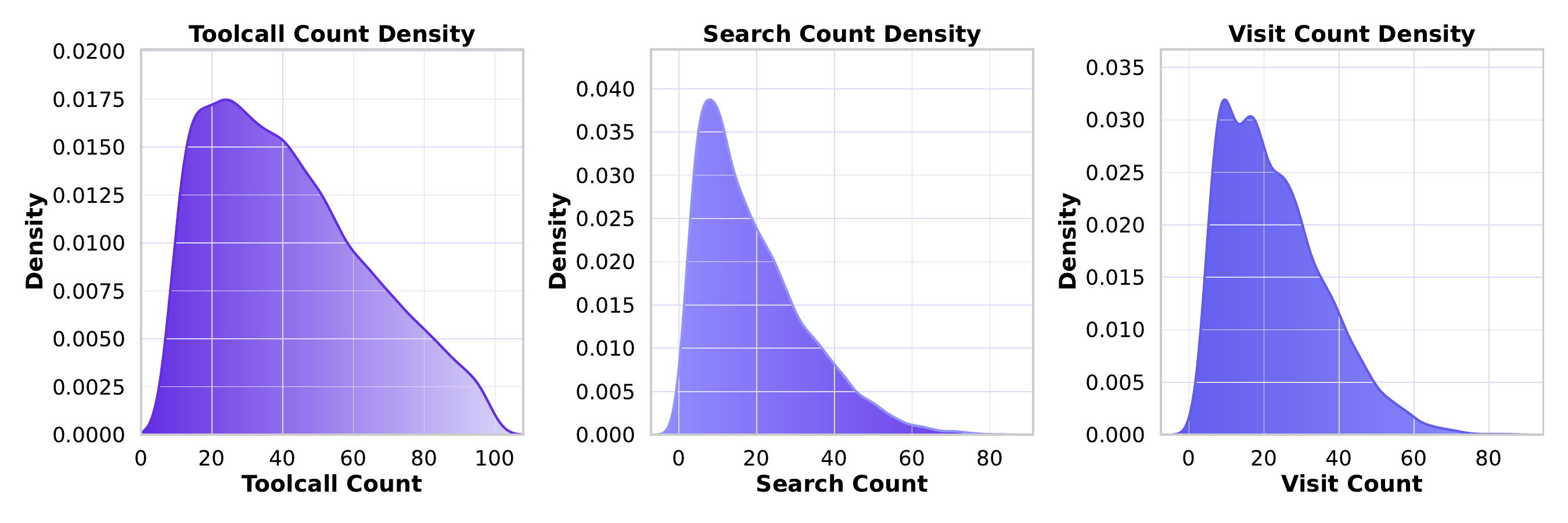}
\caption{Distribution of \textit{Search}, \textit{Visit}, and total tool call.}
\label{fig:toolcall}
\vspace{-1em}
\end{figure}

This version introduces a challenging cognitive workflow by intentionally obfuscating the query's entry points.

\textbf{Motivation and Design:}
The \texttt{Union} method, while creating multi-source tasks, has a vulnerability: an agent could solve it with simple keyword searches for each source, bypassing deeper reasoning. \texttt{Reverse-Union} inverts the information flow, forcing an agent to first deduce a core `anchor' entity (a second-layer entity) from descriptive clues and then use that entity as a pivot to expand its search.

\textbf{Example Question:} \textit{Who are the authors from the same country as the 1980s prize-winner that wrote a novel about a group of British boys stranded on an uninhabited island, and who have also won both this reward and the Booker Prize? For each of them, what is their name, country, and the respective years they won each award?}

\textbf{Construction Process:}
The construction builds upon the unified space from Version-II with a ``reverse'' logic:
\begin{itemize}[left=0.4cm]
    \item \textbf{Source:} We use the unified information space from the Nobel and Booker prize union.
    \item \textbf{Select Anchor:} An entity at the intersection of the second layers is chosen as the ``anchor,'' e.g., \textit{William Golding}.
    \item \textbf{Obfuscate Anchor:} Instead of naming the anchor, unique descriptive clues based on its third-layer attributes are generated: ``the 1980s prize-winner'' and ``wrote a novel about... British boys...'' These clues become the `Question Entities'.
    \item \textbf{Create Union Trigger:} A third-layer attribute of the anchor, his nationality (\textit{British}), is selected as the pivot for the next stage of the query.
\end{itemize}

\textbf{Required Reasoning Process:}
To solve this task, an agent must execute a two-stage process:
\begin{itemize}[left=0.4cm]
    \item \textbf{Deduction Stage:} The agent must first resolve the descriptive clues (which are third-layer entities) to identify the second-layer anchor entity. The clues ``1980s prize-winner'' and ``novel about stranded British boys'' uniquely point to \textit{William Golding}. This inferential step is crucial.
    \item \textbf{Union Stage:} Having deduced William Golding, the agent identifies his nationality (a third-layer entity in his subtree): \textit{British}. This becomes the pivot for the main query. The agent must then find all second-layer entities who (1) share this third-layer attribute (\textit{British}) and (2) have won both the Nobel Prize and the Booker Prize. This requires filtering the unified entity space to find the final set of ``Target Entities'', which includes authors like \textit{William Golding}, \textit{Kazuo Ishiguro}, and \textit{J.M. Coetzee}.
\end{itemize}

\section{Tool Call Analysis}

As shown in Figure~\ref{fig:toolcall}, our method involves a significantly large number of actions, including \textit{Search}, \textit{Visit}, and total tool calls. The density distributions indicate that tool calls often exceed several dozen per instance, with many cases surpassing 50 actions. This high frequency of actions reflects the intensive interaction and comprehensive exploration carried out by our approach, ensuring that the method thoroughly leverages available tools to achieve optimal performance.